\documentclass[lettersize,journal]{IEEEtran}
\usepackage{algorithmic}
\usepackage{array}
\usepackage{textcomp}
\usepackage{stfloats}
\usepackage[ruled,linesnumbered]{algorithm2e}
\usepackage{url}
\usepackage{bm}
\usepackage{xcolor}
\usepackage{color}
\usepackage{lscape}
\usepackage{multirow}
\usepackage{array}
\usepackage{amsmath,amssymb,amsfonts}
\usepackage{subcaption}
\usepackage{booktabs}
\usepackage{verbatim}
\usepackage{graphicx}
\usepackage{booktabs}
\def\BibTeX{{\rm B\kern-.05em{\sc i\kern-.025em b}\kern-.08em
    T\kern-.1667em\lower.7ex\hbox{E}\kern-.125emX}}
\usepackage[hidelinks]{hyperref}
\newcommand{\blue}[1]{\textcolor{blue}{#1}}
\begin{document}
\title{A Local Information Aggregation based Multi-Agent Reinforcement Learning for Robot Swarm Dynamic Task Allocation}

\author{Yang Lv, Jinlong Lei and Peng Yi

\thanks{Manuscript created November, 2023; The paper was sponsored by the National Key Research and Development Program of China under No 2022YFA1004701,  the National Natural Science Foundation of China under Grant  No. 72271187 and No. 62373283, and partially by Shanghai Municipal Science and Technology Major Project No. 2021SHZDZX0100. 

Yang Lv is with Shanghai Research Institute for  Intelligent Autonomous Systems,  Tongji University Shanghai 200092, China, (email: 726564418@qq.com).

Jinlong Lei and Peng Yi are with the Department of Control Science and Engineering, Tongii University, Shanghai, 201804, China; The Shanghai Research Institute for Intelligent Autonomous Systems,Shanghai, 201804, China;Shanghai Institute of Intelligent Science and Technology, Tongji University 200092, China, (email: leijinlong@tongji.edu.cn; pengyi@tongji.edu.cn).}
}

\markboth{IEEE TRANSACTIONS ON NEURAL NETWORKS AND LEARNING SYSTEMS,~Vol.~, No.~, ~}%
{A Local Information Aggregation based Multi-Agent Reinforcement Learning for Robot Swarm Dynamic Task Allocation}

\maketitle

\begin{abstract}
In this paper, we explore how to optimize task allocation for robot swarms in dynamic environments, emphasizing the necessity of formulating robust, flexible, and scalable strategies for robot cooperation. We introduce a novel framework using a decentralized partially observable Markov decision process (Dec\_POMDP), specifically designed for distributed robot swarm networks. At the core of our methodology is the Local Information Aggregation Multi-Agent Deep Deterministic Policy Gradient (LIA\_MADDPG) algorithm, which merges centralized training with distributed execution (CTDE). During the centralized training phase, a local information aggregation (LIA) module is meticulously designed to gather critical data from neighboring robots, enhancing decision-making efficiency. In the distributed execution phase, a strategy improvement method is proposed to dynamically adjust task allocation based on changing and partially observable environmental conditions. \blue{Our empirical evaluations show that the LIA module can be seamlessly integrated into various CTDE-based MARL methods, significantly enhancing their performance. Additionally, by comparing LIA\_MADDPG with six conventional reinforcement learning algorithms and a heuristic algorithm, we demonstrate its superior scalability, rapid adaptation to environmental changes, and ability to maintain both stability and convergence speed.} These results underscore LIA\_MADDPG’s outstanding performance and its potential to significantly improve dynamic task allocation in robot swarms through enhanced local collaboration and adaptive strategy execution. 
\end{abstract}


\section{Introduction}
\IEEEPARstart{W}{ith} the continuous advancement of modern technology, robot swarms have emerged as a significant research area, adept at handling complex tasks such as UAV swarms \cite{dong2018time}, \textit{adhoc} network relay \cite{abdulhae2022cluster}, and cooperative tracking control \cite{chen2013cooperative}. These swarms, comprising numerous small robots, excel in cooperative collaboration, underscoring the potential of collective intelligence \cite{campion2018uav}. However, efficiently coordinating these swarms for large-scale, complex tasks presents considerable challenges. A primary hurdle is the task allocation problem, which involves intelligently distributing tasks among the robots to optimize performance \cite{poudel2022task}. This issue is crucial in robotics and has broader implications for fields like industrial automation \cite{wang2022privacy}, emergency rescue \cite{chen2022trajectory}, and environmental monitoring \cite{gao2022uav}. Thus, the study of task allocation in large-scale robot swarms has become a focal point for both academic and industrial communities.

Task allocation in dynamic environments poses significant challenges in robotics. Existing research primarily focuses on scenarios involving unexpected events, such as the sudden addition or removal of robots or tasks. These changes typically occur in contexts where they are manageable and infrequent, which allows for only periodic adjustments \cite{liu2024distributed,zhang2020dynamic}. However, some real-world situations may demand a wider variety of tasks and more frequent changes, necessitating continual adaptation \cite{peng2021review}. Tasks may range from simple data collection, requiring only linear movement, to complex environmental monitoring that necessitates intricate pathways and variable speeds for effective coverage. The inherent variability of these tasks creates a strongly dynamic environment that requires consistent reallocation efforts by the robots. Furthermore, as the size of the swarms increases, the complexity of the task allocation process also escalates due to the expanded search space \cite{khamis2015multi}, posing challenges for timely responses. This study aims to address these challenges by proposing new communication protocols and coordination mechanisms to effectively manage task allocation in large-scale dynamically changing environments.

To address the challenges of task allocation, typical planning methods are categorized into centralized and distributed approaches. Centralized methods rely on a central planning system that collects all task information and uses various algorithms to devise task assignment strategies for each robot \cite{wang2012optimal,an2013simultaneous,issac2019investigations}. However, the dynamic nature of real-world tasks makes a one-time, global task allocation impractical \cite{stirling2001conditional}. Consequently, researchers have shifted towards dynamic task allocation methods that involve periodic re-planning to adapt to changing conditions and environmental dynamics \cite{lopez2019solutions,amorim2020assessing}. Nevertheless, the real-time execution of centralized planning algorithms can be time-consuming and complex, especially with a large number of robots.

Distributed task allocation methods generally offer higher computational efficiency \cite{zavlanos2008dynamic,panagou2019decentralized,yu2014target}. These methods adapt their objectives based on the communication dynamics among agents. Prominent algorithms used include the auction algorithm \cite{bertsekas1981new} and the contract network algorithm \cite{smith1980contract}, along with their various derivative algorithms \cite{luo2015distributed,lee2014resource,zhang2019task}, which have found extensive application in robot team task assignments. Although these algorithms are effective for the collaboration of multiple robots, their efficiency decreases in large-scale operations. To overcome this, a distributed autonomous decision framework based on game theory was introduced, which assumes a stable communication topology among robots and integrates social inhibition mechanisms to ensure efficient convergence to a Nash equilibrium allocation within polynomial time \cite{jang2018anonymous}. However, in a strongly dynamic multi-robot systems where communication topologies can change, efficient robot communication and collaboration become crucial. For example, in emergency rescue scenarios \cite{akgun2022using}, the status information of robots and target tasks may evolve, necessitating efficient real-time dynamic task allocation. However, existing distributed algorithms often struggle to manage these complexities effectively.

Recent studies have showcased that multi-agent reinforcement learning (MARL) methods such as MAAC \cite{iqbal2019actor}, QMIX \cite{chen2021nqmix}, and MAPPO \cite{yu2022surprising} are potent tools for tackling dynamic task allocation challenges in complex scenarios. Originating with \cite{gabel2008adaptive}, a foundational tabular-based multi-agent Q-learning framework was developed to manage dynamic tasks in unpredictable environments. This approach was further advanced by \cite{waschneck2018deep,waschneck2018optimization} through the integration of deep neural networks, enhancing the system’s adaptability. However, challenges such as environmental non-stationarity persisted. To resolve it,  \cite{qu2019dynamic} introduced a multi-agent actor-critic (MAAC) method, which significantly expedited the convergence of task allocation processes in manufacturing systems via expert-guided strategies. To further advance the robustness of  MARL, \cite{liu2020actor} introduced a parallel training mechanism employing MADDPG with asynchronous updates to better manage uncertainties in dynamic environments, marking a pivotal advancement towards more robust and scalable MARL applications. Building on this robust training foundation, \cite{luo2021real} tailored an enhanced MAPPO method specifically for dynamic multi-objective task allocation in manufacturing settings, thereby broadening MARL's practical applicability. Concurrently, \cite{wang2022solving} tackled resource contention—a frequent challenge in dynamic settings—by employing the QMIX algorithm, ensuring that agents not only met individual objectives but also contributed to achieving collective goals.

Despite these advancements, previous studies have primarily focused on the dynamics of unexpected events without fully addressing scenarios where task locations continuously change. Furthermore, while CTDE-based frameworks have utilized advanced global evaluation strategies to enhance agent coordination and policy learning, they confront a significant scalability challenge known as dimensionality explosion. This issue becomes particularly severe in large-scale tasks like robot swarm task allocation, where the state and action spaces grow exponentially as more agents are added. This exponential growth significantly increases computational complexity and memory demands. Moreover, these frameworks require each agent to process a substantial amount of information, including data from other agents. This information overload can hinder agents from focusing on data pertinent to their own decision-making, thereby reducing the overall efficiency of decision-making.

To address the challenges of task allocation in large-scale robot swarms within strongly dynamic environments, our framework adopts a local information aggregation-based Multi-Agent Reinforcement Learning (MARL) strategy. We treat each robot as an intelligent agent and transform the problem into a multi-robot cooperative system. By utilizing MARL, we develop an end-to-end solution that translates raw state data into task allocation strategies without relying on traditional plan-based methods. The MARL agents continuously refine their strategies by engaging with the environment and interacting with other robots. This dynamic interaction enables real-time and efficient collaboration among robots, allowing for swift adaptation to changing environmental conditions. Our method focuses on aggregating essential information from a subset of locally relevant agents rather than all agents, substantially reducing the dimensionality of the input space.

Building on this approach, we introduce a novel distributed MARL framework called LIA\_MADDPG, which emphasizes distributed cooperation. This framework includes a Local Information Aggregation (LIA) module, enabling each robot to manage task-related information and actively engage in information exchange and collaboration with neighboring robots. The main contributions of this paper are as follows:

\begin{itemize}
\item We investigate a new task allocation problem for robot swarms in dynamic task environments, and  set up a distributed \blue{perception and communication} model within the robot swarm network. Subsequently,  we reformulate the problem as Dec\_POMDP, that enables robots to make informed and dynamic task allocation decisions under conditions of limited information exchange  from nearby robots. 
\item  We  then propose  a novel multi-agent reinforcement learning algorithm, called LIA\_MADDPG,  that employs a combination of centralized training and distributed execution. During the centralized training phase, a local information aggregation module is incorporated to encourage robots to focus more on information beneficial to themselves during the training process. In the distributed execution phase,  robots must adapt to constantly changing conditions and make decisions based on incomplete information in dynamic environments. Thus, we  have developed a optimization method  to improve the quality of allocation policy. The method involves the analysis of observable information from each robot, allowing them to enhance allocation strategies through self-exploration and mutual collaboration. 
\item Finally, we conducted numerical experiments to compare the proposed method with \blue{six} reinforcement learning algorithms and a heuristic algorithm. \blue{In addition,
we have incorporated LIA module with two other MARL methods (e.g.,MAAC and MAPPO).} The numerical results demonstrate \blue{the broad applicability of the LIA module and highlight} the superiority of our method in terms of convergence speed, stability, and scalability, especially in tasks involving a larger number of robots. 
\end{itemize}

The  paper is organized as follows: Section \ref{section2} offers a comprehensive investigation of the robot swarm dynamic task allocation problem, along with its  reformulation as a Dec\_POMDP model. Section \ref{section3}  provides  an in-depth description of the proposed method LIA\_MADDPG. Section \ref{section4} presents simulation results and the associated discussions. Lastly, some concluding remarks are given in Section \ref{section5}.   

\section{Problem Descriptions}\label{section2}
In this section, we begin by delving into the details of modelling the robot swarm task allocation problem in dynamic task environments. Subsequently, we reformulate this problem as a Dec\_POMDP. 

\subsection{Robot Swarm Task Allocation Problem }\label{subsection2.1}
The robotic swarm task allocation problem  presents a challenging scenario where a swarm of available robots needs to be allocated to mobile tasks in dynamic environments. The primary objective  is to efficiently assign robots to tasks  according to  some specific performance metric. This problem  comprises three essential components. The first part focuses on the mobile task set and their corresponding motion models. The second part addresses the robot swarm network and provides a detailed description of the associated  distributed communication model. The last part designs an appropriate  performance metric and introduces the  optimization model. 

\paragraph{\textbf{Movable Task Set}} Define $\mathcal{M}=\{1,2,...,M\}$ as the set of movable tasks within the time series ${\mathcal{T}}:=\{0,1,\ldots, T\}$. Let $v_j^m$ represent the speed of task $j\in\mathcal{M}$, and $\theta_j^{m,t}$ denote the movement angle of task $j$ at time  $t$. Let $P_j^{m,t}=(x_j^{m,t},y_j^{m,t})$  represent the two-dimensional spatial coordinates of task $j$ at time $t$. The motion state of the tasks is determined by their speed and movement angle and can be updated using \eqref{eq1}.
\begin{equation}
P_j^{m,t+1}=\left\{\begin{matrix}x_j^{m,t}+v_j^m\tau cos(\theta_j^{m,t})\\ y_j^{m,t}+v_j^m\tau sin(\theta_j^{m,t})\end{matrix}\right.\label{eq1}
\end{equation}
where $\tau$ represents the decision time step.

Each task in $\mathcal{M}$ requires a substantial allocation of robots while maintaining a moderate demand, which means it has two characteristics: (1) each task necessitates the allocation of multiple robots for execution, and (2) there is a limitation on the number of robots required per task. Therefore, we define ${\bar{h}}_j$ as the maximum number of robots that task $j\in\mathcal{M}$ can accommodate.

\begin{figure}[ht]
\centering
\includegraphics[width=0.47\textwidth]{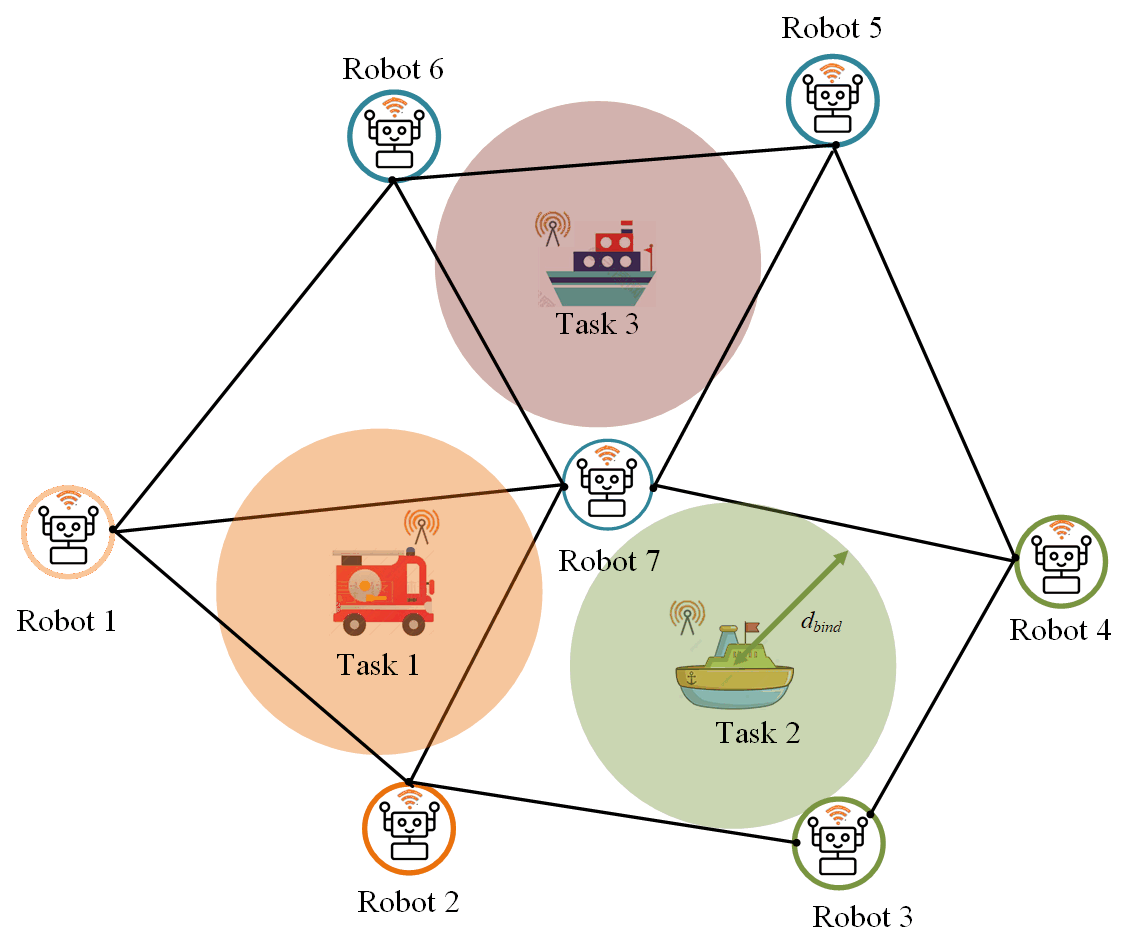}
\caption{Distributed \blue{perception and communication}  in Robot Swarm Network}
\label{swarm_net}
\end{figure}

\paragraph{\textbf{Robot Swarm Network}} Define the robot swarm network as $\mathcal{G}=\{\mathcal{N},\mathcal{E}\}$, where $\mathcal{N}=\{1,2,\ldots,N\}$ represents the node set of robots, and $\mathcal{E}\subset\mathcal{N}\times\mathcal{N}$ represents the \blue{perception}  relationships among robots. Each robot corresponds to a node $\mathcal{V}$ in the network. For an edge $(i,i')\in\mathcal{E}$, it signifies that robot $i$ can \blue{observe} information from robot $i'$. Therefore, we can denote the set of all neighboring robots of robot $i$ as  $\mathcal{N}_i\triangleq\{i'\in\mathcal{N}:(i,i')\in\mathcal{E}\}$.
We assume that each robot   $i\in \mathcal{N}$ is a rational agent with limited \blue{perception and communication} capabilities, which  can instantly access the status of all tasks and information from neighboring robots $i'{\in\mathcal{N}}_i$. Based on the \blue{observed and received}  information and the current environment,  robot   $i\in \mathcal{N}$ can dynamically choose target tasks, and  subsequently, moves at a constant velocity $v_i^r$ towards the chosen task.  Fig.\ref{swarm_net} depicts a example in which seven robots are interconnected, forming a distributed \blue{perception and communication} network. \blue{It should be noted that the interaction between the robot and the task occurs through  data exchange}. This network enables the robots to observe and collaborate with neighbors by sharing observation data, task status, and decision outcomes. 

We define $g_i^t\in\mathcal{M}$ as the target task of robot $i\in\mathcal{N}$ at time $t\in\mathcal{T}$, and $d_{i,g_i^t}^t\in\mathbb{R}$ as the Euclidean distance between robot $i$ and its target task $g_i^t$ at the $t$-th decision moment. Let  $d_{bind}>0$ denote the predefined association distance,  and $T_i\in\mathcal{T}$ be the finish time of robot $i$. If at some time $t$, robot $i$'s distance \blue{from its target task satisfies}  $d_{i,g_i^{t}}^t\le d_{bind}$, then robot $i\in\mathcal{N}$ becomes bound to this task $d_{i,g_i^{t}}^t\in\mathcal{M}$, and $T_i\triangleq t$ is set as the binding time. At time $T_i$, the robot is considered to have completed the task assignment and it cannot change its target task anymore. Let $P_i^{r,t+1}=\left(x_i^{r,t},y_i^{r,t}\right)$ represent the coordinates of robot $i\in\mathcal{N}$,   and $\lambda_{i,g_i^t}^{t+1}=\frac{y_{g_i^t}^{m,t}-y_i^{r,t}}{x_{g_i^t}^{m,t}-x_i^{r,t}}$ represent the direction vector of robot $i\in\mathcal{N}$. Then each robot moves towards  the chosen target task and can update their motion state according \eqref{eq2}. It should be emphasized that once the robot is bound, its motion state will consistently synchronize with its target task. 
\begin{equation}
P_i^{r,t+1}=\left\{\begin{matrix}x_i^{r,t}+v_i^r\tau cos\left(\lambda_{i,g_i^t}^{t+1}\right)\\y_i^{r,t}+v_i^r\tau sin\left(\lambda_{i,g_i^t}^{t+1}\right)\\\end{matrix}\right. ,\label{eq2}
\end{equation}

\paragraph{\textbf{Utility Function}}For each robot $i\in\mathcal{N}$, let
 $u_i$ represent the utility function that robot $i\in\mathcal{N}$ can obtain when assigned to some task  $j\in\mathcal{M}$. We design it to comprise the task feedback rewards $u_i^1$ and movement cost reward $u_i^2$. In the following, we will separately introduce them.

The first item $u_i^1$ represents the reward that the robot receives upon completing a task assignment, and the assigned task is denoted  as $j=g_i^{T_i}$ for convenience. Define $h_j^t$ as the number of robots already bounded to task $j\in\mathcal{M}$ at time $t\in\mathcal{T}$, and $r_{i,j}$ as the reward that robot $i\in\mathcal{N}$ can obtain from task $j\in\mathcal{M}$. In fact, only when the number of robots bounded to task $j$ is less than  ${\bar{h}}_j$ (the maximum number of robots that task $j\in\mathcal{M}$ can accommodate), robot $i\in\mathcal{N}$ can receive rewards from task $j\in\mathcal{M}$. Therefore, the expression for $u_i^1$ is as follows.
\begin{equation}
u_i^1 =
\begin{cases}
    \begin{aligned}
        & r_{i,j}, && \text{if}\enspace h_j^{T_i}<\bar{h}_j\\
        & 0, &&\text{otherwise}.
    \end{aligned}
\end{cases}\label{eq3}
\end{equation}

The second component is the cost incurred by robot $i\in\mathcal{N}$ while moving towards the target task. It is denoted as $u_i^2=\sum_{t=1}^{T_i}{v_i^r\tau}$.

Therefore, for each robot $i\in\mathcal{N}$  with the assigned task  denoted  as $j=g_i^{T_i}$, its utility function can be described below.
\begin{equation}
u_i=u_i^1-u_i^2=
\begin{cases}
    \begin{aligned}
        & r_{i,j}-\sum_{t=1}^{T_i}v_i^r\tau, && \text{if}\enspace h_j^{T_i}<\bar{h}_j\\
        & -\sum_{t=1}^{T_i}v_i^r\tau, &&\text{otherwise}.
    \end{aligned}
\end{cases}\label{eq4}
\end{equation}

\paragraph{\textbf{Optimization Problem Formulation}} Our objective is to dynamically select target task for each robot   based on the states of  neighboring robots and target tasks, with the aim of maximizing the cumulative utility of all robots.  To achieve this, we formulate the aforementioned problem as a collaborative optimization problem consisting of three key components: decision variables, an objective function, and constraints.

\begin{figure}[h]
\centering
\includegraphics[width=0.47\textwidth]{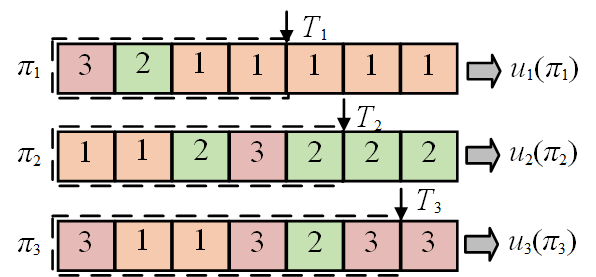}
\caption{ A simple diagram of decision variable based on the case in Fig.\ref{swarm_net}}
\label{decision_v}
\end{figure}

We define the decision variable for robot $i\in\mathcal{N}$ as $\pi_i$, which represents the target task selected by the robot throughout the time series $\mathcal{T}$, i.e., $\pi_i=\{g_i^1,g_i^2,\dots, g_i^T\}$. The utility function \eqref{eq4} is employed as the objective function to assess the value of the current decision variable $\pi_i$. A simple diagram  for the case in Fig.\ref{swarm_net} is shown in Fig.\ref{decision_v} . Additionally, we must adhere to the following constraints:

1). The length of the time series $\mathcal{T}$ is restricted to the maximum time required for the last robot to reach its target task, i.e., $T=|\mathcal{T}|=\max_{i\in\mathcal{N}}T_i$, where $d_{i,g_i^{T_i}}^t\leq d_{\text{bind}}$.

2). Once robot $i\in\mathcal{N}$ is bound to its target task at time $T_i$, it cannot switch to another target task, i.e., $g_i^t= g_i^{T_i} $ for any $ t\in[T_i,T]$.

3). Let $x_{i,j}\left(t\right)\in \{0,1\}$ be a binary decision variable indicating whether robot $i\in\mathcal{N}$ chooses task $j\in\mathcal{M}$ at time $t$,
where $x_{i,j}\left(t\right)=1$ signifies  that  task $j\in\mathcal{M}$ is chosen by   robot $i\in\mathcal{N}$ at time $t$. Each robot  can only select one target task at each time $t$, hence  $\sum\limits_{  j\in\mathcal{M}}x_{i,j}(t)= 1,\forall i\in\mathcal{N},\forall t\in\mathcal{T}$. 

Based on the above discussions, the robot swarm task allocation problem in dynamic task settings can be formulated as the  following   optimization problem.
\begin{equation}
\begin{cases}
    \max\limits_{\pi_i}\sum\limits_{i\in\mathcal{N}}u_i(\pi_i)\\
    \text{s.t.}\enspace 
    \begin{aligned}
        & T=|\mathcal{T}|=\max_{i\in\mathcal{N}}T_i ,~ d_{i,g_i^{T_i}}^t\leq d_{\text{bind}};\\
        & g_i^t= g_i^{T_i},~\forall t\in[T_i,T] ;\\
        &\sum\limits_{  j\in\mathcal{M}}x_{ij}(t)= 1,\forall i\in\mathcal{N},\forall t\in\mathcal{T};\\
        & x_{i,j}(t)=
            \begin{cases}
                \begin{aligned}
                    1,j=g_{i}^t\\
                    0,j\neq g_{i}^t\\
                \end{aligned}
            \end{cases}
    \end{aligned}
\end{cases}\label{eq5}
\end{equation}

\subsection{Reformulation as a  Decentralized Partially Observable Markov Decision Process}\label{subsection2.2}
In the aforementioned robot swarm task allocation problem, robots are responsible for selecting their target tasks based on limited observations. We then reformaute the problem as a Dec\_POMDP \cite{oliehoek2016concise}, which comprises the tuple $\mathcal{H}=\{\mathcal{S},\mathcal{A},\mathcal{P},\mathcal{Z},\mathcal{O},\mathcal{R}\}$. Specifically, given the current state $s^t\in\mathcal{S}$, each robot $i\in\mathcal{N}$ can obtain its local observations $o_i^t\in\mathcal{O}$ through the observation function $\mathcal{Z}$. Subsequently, it takes actions $a_i^t\in\mathcal{A}$ according to a predetermined policy and receives corresponding rewards $r_i^t\in\mathcal{R}$. The environment then transitions to the next state based on the state transition function $\mathcal{P}$. Next, we will provide a detailed explanation of the Dec\_POMDP model.

\paragraph{\textbf{State Space}} The state information of robot  $i\in\mathcal{N}$ at time $t$ is denoted as $s_i^{r,t}=\{x_i^{r,t},y_i^{r,t},v_i^r,\lambda_i^t,c_i^t\}$. Here, $c_i^t\in\{0,1\}$ represents the working status of robot $i$, with  $c_i^t=0$ indicating that the robot is already bound to a task, and  $c_i^t=1$ indicating that the robot can actively choose a task. The state information of task $j\in\mathcal{M}$ is defined as $s_j^{m,t}=\{x_j^{m,t},y_j^{m,t},v_j^m,\theta_j^{m,t},h_j^{m,t}\}$. These symbols are consistent with the previous text. Thus, the state space $\mathcal{S}$ consists of two components: the robots state space $\mathcal{S}^r=\{s_1^{r,t},s_2^{r,t},\ldots,s_N^{r,t}\}$ and the tasks state space  $\mathcal{S}^m=\{s_1^{m,t},s_2^{m,t},\ldots,s_M^{m,t}\}$.

\paragraph{\textbf{Action Space}} The action of robot $i$ involves selecting its desired target task $g_i^t$ at each decision moment. We utilize a M-dimensional one-hot encoded vector  as robot $i$'s action $a_i^t$. If robot $i$ selects task $j=g_i^t$ at time $t$, then the $j$-th element of action $a_i^t$ is set to 1, while all other elements are set to 0. Additionally, if the control information of robot $i$ $c_i^t=0$, i.e., bound to the target task, then $a_i^{t+1}=a_i^t$. Thus, the joint action space of all robots is defined as  $\mathcal{A}=\{a_{1}^{t},a_{2}^{t},\ldots,a_{N}^{t}\}$.

\paragraph{\textbf{State Transition Function}}$\mathcal{P}(\mathcal{S},\mathcal{A})\colon\mathcal{S}\times\mathcal{A}\to\mathcal{S}'$ is employed to describe the state transition of the environment after robots execute joint actions. Similarly, the state transition function encompasses both the task state transition function $\mathcal{P}_i^m$ and the robot state transition function $\mathcal{P}_i^r$ .

The task state transition function primarily involves updating the task's position according to \eqref {eq1}.  Furthermore, based on the actions taken by the robots, it is necessary to update the number of robots assigned to the task $j\in\mathcal{M}$, i.e., $h_j^{t+1}=h_j^{t}+num_j$, where $num_j$ signifies the count of robots assigned to task $j$ at time $t$. Therefore, the task state transition function $\mathcal{P}_j^m$ for task $j\in\mathcal{M}$ is as follows:
\begin{equation}
        \begin{aligned} 
            s_j^{m,t+1}=
            & \{x_j^{m,t}+v_j^m\tau cos(\theta_j^{m,t}),y_j^{m,t}+v_j^m\tau sin(\theta_j^{m,t}),\\
            & v_j^m,\theta_j^{m,t+1},h_j^{t+1}\}\label{eqn}\\
        \end{aligned}
\end{equation}

The robot state transition function is also a component of the environmental state transition. Since each robot is independent, the state transition function $\mathcal{P}_i^r$ for robot $i$ depends solely on the its current state $s_i^t$ and the action $a_i^t$. It can be categorized into three scenarios. 

 1). $c_i^t=0$, i.e., the robot has been bound to the target task, and  then synchronized their motion state with its target tasks.

2). $c_i^t=1\overset{a_i^t}{\to}c_i^{t+1}=1$, i.e., the robot remains unbound to any task after executing the action. Then the state transition  involves updating the robot's position according to \eqref {eq2}. 

3). $c_i^t=1\overset{a_i^t}{\to}c_i^{t+1}=0$, i.e., the robot becomes bound to a target task after executing the action. In this case, besides updating the robot's position, it is vital to update the robot's control information $c_i^t$ as well.

Hence, the transition function $\mathcal{P}_i^r$ for robot $i\in\mathcal{N}$ after executing action $a_i^t$ can be expressed as follows:
\begin{equation}
s_i^{r,t+1}=
    \begin{cases}
        \begin{aligned}
            & \{x_{g_i^t}^{m,t+1},y_{g_i^t}^{m,t+1},v_{g_i^t}^m,\theta_{g_i^t}^{m,t+1},0\},  if\enspace c_i^{t}=0,\\
            & \{x_i^{r,t+1},y_i^{r,t+1},v_i^r,\lambda_i^{t+1},1\},  if\enspace c_i^{t}=1,c_i^{t+1}=1,\\
            & \{x_i^{r,t+1},y_i^{r,t+1}, v_i^r,\lambda_i^{t+1},0\},  if\enspace   c_i^{t}=1,c_i^{t+1}=0.\\
        \end{aligned}
    \end{cases}\label{eq6}
\end{equation}

\paragraph{\textbf{Observation Space}}  In the above-mentioned model,  each robot has limited \blue{perception and communication} capabilities that can \blue{obtain} information with each task and neighboring robots. We define $\mathcal{N}_i^t=\{na_1^t,na_2^t,\ldots,na_{\alpha_i}^t\}$ as the set of neighboring robots of robot $i\in\mathcal{N}$ at time $t\in\mathcal{T}$, where $\alpha_i$ represents the number of neighboring robots, depending on the \blue{perception} capacity of robot $i$. Therefore,   each robot  $i\in \mathcal{N}$  can only observe a portion of the state, and its
  local observation information  primarily includes three parts. 1). Self-state information $o_{Self}^{t}=s_{i}^{r,t}$. 2). The relative state information between robot $i$ and all tasks $o_{Task}^t=\{\Delta x_{i,j}^t,\Delta y_{i,j}^t,\Delta v_{i,j}^t,\Delta\theta_{i,j}^t,h_j^t,\kappa_{i,j},\forall j\in\mathcal{M}\}$, where $\Delta x_{i,j}^t$, $\Delta y_{i,j}^t$, $\Delta v_{i,j}^t$, and $\Delta\theta_{i,j}^t$  represent the horizontal and vertical coordinates, relative velocity and relative motion direction between robot $i$ and task $j$, respectively, $h_{j}^{t}$ denotes the number of robots already bound to task $j\in\mathcal{M}$ at time $t\in\mathcal{T}$, and $\kappa_{i,j}$ represents the normalized weight of the reward that robot can obtain from target task. 3) The relative state information between robot $i\in\mathcal{N}$ and its neighboring ones $o_{Neighbor}^t=\{\Delta x_{i,i'}^t,\Delta y_{i,i'}^t,\Delta v_{i,i'}^t,\Delta\theta_{i,i'}^t,g_{i'}^{t-1},\forall i'\in \mathcal{N}_i^t\}$. Therefore, the local observation value for robot $i\in\mathcal{N}$ can be defined as $o_t^i=\{o_{Self}^t,o_{Task}^t,o_{Neighbor}^t\}$.

\paragraph{\textbf{Reward Function}} Rewards play a crucial role in reflecting the environment's response to changes in the agents' states caused by their actions. However, in the case of dynamic task allocation problems with sparse rewards, designing effective learning strategies becomes complex and challenging. This is because when rewards are not observable, the policy gradient becomes zero, making it impossible to improve the policy. To tackle this issue, one approach is to leverage prior knowledge of the problem and propose a non-sparse reward function  that can provide more informative feedback to guide the learning process.

Based on the optimization model in \eqref{eq5} and the limited demand constraints of tasks, we will design  a reward function consisting of three components as follows.
\begin{equation}
    r_t^i=r_t^{i,dis}+r_t^{i,step}+r_t^{i,final},\label{eq7}
\end{equation}
where  the first two items are designed to be non-sparse, and the last term represents the ultimate reward obtained when a robot reaches a target task.

 In details, $r_{t}^{i,dis}=-\varphi_{2}$ is a constant negative reward given to robot  $i\in \mathcal{N}$ at each time step until  reaching its target task. $r_t^{i,step}$ is designed based on the utility function $u_i^1$ and aims at guiding robots to satisfy the limited demand constraints of tasks. It is defined as: 
\begin{equation}
    r_t^{i,step}=\varphi_3\left(\bar{\hbar}_j-\hbar_j^t\right).\label{eq8}
\end{equation}
The above reward mechanism requires robots to consider the task's demand when choosing a target. If the number of robots bound to task $j\in\mathcal{M}$ exceeds the task's maximum demand at time $t\in\mathcal{T}$, these additional robots will be penalized. Moreover, as the number of robots exceeding the maximum demand increases, the penalty strength will also  increase to prevent excessive concentration of robots on a single target. 

The final term $r_t^{i,final}$ represents the ultimate reward obtained when a robot reaches a target task and is defined by
\begin{equation}
r^{i,final}_t=
\begin{cases}
    \begin{aligned}
        & \varphi_1r_{i,j}, && \text{if}\enspace h_j^t<\bar{h}_j,\\
        & 0, &&\text{otherwise}.
    \end{aligned}
\end{cases}\label{eq9}
\end{equation}

 Additionally, it is crucial to address an exceptional scenario where robots frequently switch target tasks without making progress towards completion. Therefore, it is necessary to take a balance  between $\varphi_1$ and $\varphi_2$ to ensure that robots are still influenced by the penalty term  $r_{t}^{i,dis}$  when frequent target switching occurs.

\section{A Novel MARL Algorithm with Local Information Aggregation}\label{section3}
In this section, we systematically introduce our novel MARL algorithm with local information aggregation, including the main components of the algorithm and the design inspiration behind them.

\subsection{\blue{Key Modules and Mechanisms of LIA\_MADDPG}}\label{subsection a}
MADDPG is a classic multi-agent deep reinforcement learning algorithm  to address multi-agent problems in mixed cooperative-competitive environments \cite{lowe2017multi}. However, when applied to large-scale problems like the robot swarm task allocation problem considered in this work, MADDPG encounters difficulties in coordinating learning due to scalability issues. To overcome this challenge, we introduce a novel distributed method called LIA\_MADDPG, which incorporates the Local Information Aggregation (LIA) module. 

In a typical MADDPG setup with $N$ agents, it is  necessary for all robots  to maintain a set of policy networks $\left\{\mu_i\right\}_{i=1}^N$ and their corresponding target policy networks $\{\mu_{i}^{\prime}\}_{i=1}^{N}$ , as well as a set of value networks $\left\{Q_i\right\}_{i=1}^N$ and their target value networks $\{Q_{i}^{\prime}\}_{i=1}^{N}$. However, for the large-scale homogeneous robots in the robot swarm task allocation problem, a policy-sharing approach can be employed during training to simplify the network structure. In this approach, all robots share a common policy network  $\bar{\mu}$ with parameters $\theta^a$ and a shared value network  $\bar{Q}$ with parameters $\theta^q$. Additionally, the shared policy network $\bar{\mu}$ also has a corresponding target policy network $\bar{\mu}'$ with parameters $\theta'^a$, and the shared target value network $\bar{Q}'$ with parameters $\theta'^q$. This policy-sharing approach reduces the number of networks required, resulting in a simpler algorithm with lower computational complexity.

\paragraph{\textbf{Shared Policy Network}}
The shared policy network $\bar{\mu}$ utilizes fully connected layers to handle partial observability and sequential information from the environment. It incorporates residual connections \cite{he2016deep} and batch normalization \cite{ioffe2015batch} to enhance the performance and training efficiency of deep neural networks. For  each robot $i\in \mathcal{N}$, $\bar{\mu}$ takes the current time-step observations $o_i^t$ as input and generates the corresponding action $a_i^t$ for the next time-step.
\begin{equation}
    a_{i}^{t}=\bar{\mu}(o_{i}^{t};\theta^{a}).\label{eq10}
\end{equation}
Similarly, the action output based on the target network can be defined as:
\begin{equation}
    a_i^{t+1}=\bar{\mu}'(o_i^{t+1};\theta'^{a}).\label{eq11}
\end{equation}

\paragraph{\textbf{Local Information Aggregation (LIA)}}
The value network of the traditional MADDPG algorithm takes the joint observations $\boldsymbol{o}^t = (o_1^t, o_2^t, ..., o_N^t)$ and joint actions $\boldsymbol{a}^t = (a_1^t, a_2^t, ..., a_N^t)$ from all robots to compute the \textit{Q}-value $\bar{Q}(\boldsymbol{o}^t, \boldsymbol{a}^t, \theta^{q})$. While this approach mitigates the impact of non-stationary environments by considering all agents' states, \blue{meanwhile} it faces exponential growth in input dimensions as the number of agents increases, making state evaluation difficult. In our problem, a robot $i$ only needs to consider the states of certain robots, referred to as “locally related robots.” Therefore, during centralized training, we focus only on information from these locally related robots. This selective focus helps mitigate the challenges of dimensionality but introduces another issue: the number of locally related robots is uncertain, leading to variable input dimensions for the value network. To address these challenges, we propose a Local Information Aggregation (LIA) module, which aims to solve two key problems: (1) defining and selecting the set of locally related robots, and (2) handling the variability in the number of these related robots, which leads to issues with the input dimensions of the value network. Next, we will introduce the detailed definition of locally related robots.

In our study, we assume that robots performing the same action in close proximity (referred to as “locally related robots”) collectively influence their environment. This assumption is crucial for understanding group dynamics. The synchronized efforts of these robots significantly amplify their impact. For example, when multiple robots gather at a specific location or perform the same action (such as selecting the same target task), their interactions due to spatial proximity can influence each other's decisions. Similarly, robots that choose the same task will also affect each other because of the limited capacity of each task.

To accurately capture this phenomenon, we define locally related robots ($\mathcal{G}_{i}^t$) not only through spatial proximity but also through action synchronization. Specifically, $\mathcal{G}_{i}^t=\{{\mathcal{N}}_{i}^t,{\mathcal{L}}_{i}^t\}$, where $\mathcal{N}_{i}^t=\{n a_{1}^t,n a_{2}^t,\ldots,n a_{\alpha_{i}^t}\}$ includes robots that are near to robot $i$, and $\mathcal{L}_{i}^t \triangleq \{i' \in \mathcal{N} : a_{i}^t = a_{i'}^t\}$ includes robots executing the same action as robot $i$. The influence of these robots on $i$'s decision-making is significant because nearby robots can create local environmental conditions that directly affect robot $i$'s operational context. Simultaneously, robots performing the same actions may lead to competition or cooperation for resources, further impacting $i$'s decisions. This dual consideration integrates collective behavior more effectively into the training process, enabling our model to accurately predict and adapt to group dynamics, thereby enhancing each robot's decision-making process. By recognizing how the same actions performed by nearby robots amplify environmental impacts, our model not only captures dynamics more precisely but also reduces computational load by focusing on a manageable subset of influential robots, simplifying the training process. Thus, the local information needed for training can be defined as follows:

\textbf{Local Information Set}: Define the local information for robot $i$ as the collection of observations and actions from its locally related robots, represented as $L_i = {(o_k^t, a_k^t) \mid k \in \mathcal{G}_i^t}$, where $o_k^t$ and $a_k^t$ represent the observation vector and the action vector of robot $k$ at time $t$.

It's evident that the dimensionality of robot $i$'s local information set vector is related to the number of its locally related robots. However, the set of locally related robots varies among different robots, and even for the same robot, this set may change over time. In such cases, concatenating robot $i$'s state information with its local information for input into the value network results in inconsistent input vector dimensions. To address this issue, inspired by the mean field RL approach from \cite{yang2018mean}, we developed a local information aggregation function, denoted as $\varphi_{i}$. The function $\varphi_{i}$ takes the local observations and actions of the robots in $\mathcal{G}_{i}^t$ as inputs and generates a fixed-dimensional aggregated information vector suitable for input into the value network. The LIA function $\varphi_{i}$ is expressed as follows:
\begin{equation}                        \varphi_i\Big(o_k^t\Big)=\sum_{k\in\mathcal{G}_i^t}w_{i,k}^to_k^t,\label{eq14}
\end{equation}
\begin{equation}                    
    \varphi_i\big(a_k^t\big)=\sum_{k\in\mathcal{G}_i^t}w_{i,k}^ta_k^t,\label{eq15}
\end{equation}
where $w_{i,k}^t$ is distance-dependent weight coefficients defined based on the distance $d_{i,k}^t$ between robot  $i\in\mathcal{N}$  and its locally relevant robot $k\in\mathcal{G}_{i}^t$. These weight coefficients are assigned to each locally relevant robot $k\in\mathcal{G}_{i}^t$ .
\begin{equation}                    
    w_{i,k}^t=\frac{\exp\left(\beta ln(d_{i,k}^t)\right)}{\sum_{k\in\mathcal{G}_i^t}\exp\left(\beta ln(d_{i,k}^t)\right)}.\label{eq16}
\end{equation}

\textbf{Remark}: The weighting method defined in \eqref{eq16} and the attention mechanism \cite{vaswani2017attention} represent two distinct methods of aggregating agent information in multi-agent systems. Eq.\eqref{eq16} calculates weight coefficients $w_{i,k}^t$ based on the physical distance $d_{i,k}^t$ between robots, using an exponential function scaled by a factor $\beta$. This deterministic approach prioritizes agents based on proximity, ideal for applications such as robotic swarms where spatial relationships are crucial. In contrast, attention mechanisms in neural networks learn to assign weights adaptively from data, allowing the model to dynamically assess the relevance of each agent based on the task and contextual interactions. While \eqref{eq16} provides a fixed, proximity-based weighting ideal for environments where agent distance is a dominant factor, attention mechanisms offer greater flexibility by adapting to complex data relationships beyond physical proximity. This adaptability makes them suitable for a wider range of applications where contextual nuances of agent interactions are critical.

By aggregating the local information from locally related robots, $\varphi_{i}$ enables robot $i$ to effectively consider the collective knowledge and actions of its nearby peers. This aggregation function not only ensures consistency in the input dimensions for the value network but also captures the intricate dynamics of the environment more precisely. By integrating observations and actions from locally related robots, $\varphi_{i}$ allows robot $i$ to make more informed decisions, reflecting both the immediate environmental conditions and the synchronized actions of its peers. Consequently, this approach enhances the model's ability to predict and adapt to group dynamics, ultimately improving the overall decision-making process and operational efficiency of the robot.

\paragraph{\textbf{Shared Value Network}}
We introduce a novel centralized value function, denoted as ${G}$. The corresponding shared value network $\bar{G}$, takes its own local observation $o_{i}^{t}$ and action $a_{i}^{t}$, as well as the aggregated information $\varphi_{i}\left(o_{k}^{t}\right)$ and $\varphi_{i}\left(a_{k}^{t}\right)$ as input. $\bar{G}$ then computes the evaluation value $q_i^{t}$ of robot $i$ at time \textit{t}.
\begin{equation}                    
    q_i^{t}=\bar{G}[o_{i}^{t},a_{i}^{t},\varphi_{i}(o_{k}^{t}),\varphi_{i}(a_{k}^{t}) {;}\theta^{g}], k\in{\cal G}_{i}^{t},k\neq i.\label{eq17}
\end{equation}
Likewise, the target shared value network $\bar{G}'$ follows the same structure.
\begin{equation}            q_i'^{(t+1)}=\bar{G}'[o_i^{t+1},a_i^{t+1},\varphi_i(o_k^{t+1}),\varphi_i(a_k^{t+1})\blue{;}\theta'^{g}], k\in{\cal G}_{i},k\neq i.\label{eq18}
\end{equation}

\paragraph{\textbf{Network Updating}}
In multi-agent reinforcement learning, a critical challenge during the network update process is the environmental non-stationarity arising from the policy changes of other agents during training.  Take  Ind\_DDPG \cite{lillicrap2015continuous} as an example. It learns independently for each agent by treating other agents as components of the environment and evaluates the $Q$ function according to the following.
\begin{equation*} 
    \begin{aligned}
        Q^*(\pmb s^t,& a_i^t)= \\
        &\sum_{\pmb s^{t+1}}p(\pmb s^{t+1} \mid \pmb s^t,a_i^t)[r_i^{t+1}+\gamma\max_{a_i^{t+1}}Q_i^*(\pmb s^{t+1},a_i^{t+1})],\\
    \end{aligned} 
\end{equation*}
where $p(\pmb s^{t+1}\mid \pmb s^t,a_i^t)$ represents the probability of agent $i$ transitioning from state $\pmb s^t$ to the next state $\pmb s^{t+1}$ when taking action $a_i^t$. However, it is worth noting  that during the learning process, changes in other agents' strategies can introduce non-stationarity in the state transition probability $p(\pmb s^{t+1}\mid \pmb s^t,a_i^t)$. Consequently, relying solely on a temporal-difference (TD) based approach for iteratively learning the \textit{Q} function may result in suboptimal learning outcomes.

In our study, this problem can be effectively addressed by \blue{introducing} the extended Q-function $G$, which evaluates the expected total reward that robot $i\in\mathcal{N}$ can obtain based on its own local observation $o_{i}$, action $a_{i}$ and the aggregated information from locally connected robots. We then derive the Bellman optimality equation for the extended $Q$-function of robot $i$'s policy.
\begin{equation} 
    \begin{aligned}
        G^*(o_i^t,a_i^t, & \varphi_i(o_k^t),\varphi_i(a_k^t))= \\
        &\sum_{\pmb o^{t+1}}p(\pmb o^{t+1}\mid \pmb o^t,a_i^t)[r_i^{t+1}+\\
        &\gamma\max\limits_{a_i^{t+1}}G(o_i^{t+1},a_i^{t+1},\varphi_i^{t+1}(o_k),\varphi_i^{t+1}(a_k))].\\
    \end{aligned}\label{eq22}
\end{equation}
The state transition $p(\pmb{o}^{t+1}\mid\pmb{o}^{t},a_{i}^{t})$ can be decomposed as  
\begin{equation} 
    p(\pmb o^{t+1}\mid \pmb o^t,a_i^t)=\sum_{a_{-i}^t}p(a_{-i}^t\mid o_{-i}^t)p(\pmb o^{t+1}\mid o_i^t,a_i^t,a_{-i}^t),\label{eq23}
\end{equation}
where $p(\boldsymbol{a}_{-i}^t\mid\boldsymbol{o}_{-i}^t)=\prod_{j\neq i}p({a}_{j}^t\mid {o}_{j}^t)$. For robot $i\in\mathcal{N}$, the source of environmental instability arises from the strategy distributions $p\left(a_{-i}^{t}|o_{-i}^{t}\right)$ of other robots related to $i$. In this article, these relevant robots can be approximated as the local relevant robot set $\mathcal{G}_i^{t}$ of robot $i$. Therefore, the state transition in \eqref{eq23} can be redefined as 
\begin{equation} 
    \begin{aligned}
        p(\pmb o^{t+1}\mid & \pmb o_i^t,a_i^t)=\\
        &\sum_{a_k^t}p(a_k^t\mid o_k^t)p(\pmb o^{t+1}\mid o_i^t,a_i^t,a_k^t)_{k\in\mathcal{G}_i^t,k\neq i},\\
    \end{aligned}\label{eq24}
\end{equation}
where $\textbf{\textit{a}}_{k}^{t}$ and $\textbf{\textit{o}}_{k}^{t}$  represent the local agent actions and observations incorporated into the extended \textit{Q}-function through the aggregation function $\varphi_{i}$. When we know the actions taken by the relevant robots of robot $i$, even if the policies change, the environment can be considered static. This is because for any policies $\pi_i\neq\pi_i',\pi_k\neq\pi_k'$, we have $p\big(\pmb o^{t+1}\mid \pmb o^t,a_i^t,a_k^t,\pi_i,\pi_k\big)_{k\in G_i^t\cdot k\neq i}=p\Big(\boldsymbol{o}^{t+1}\bigm|\boldsymbol{o}^{t},a_{i}^{t},a_{k}^{t}\Big)=p\big(\boldsymbol{o}^{t+1}\mid\boldsymbol{o}^{t},a_{i}^{t},a_{k}^{t},\pi_{i}^{\prime},\pi_{k}^{\prime}\big)_{k\in{\cal G}_{i}^{t},k\neq i}$. As a result, we can learn the extended $Q$-function $G$ based on the above equation in a stable environment.

Therefore, the shared value network  $\bar{G}$ can be trained to approximate extended \textit{Q}-function-function $G$ by minimizing the squared TD error. The loss function is given  as follows: 
\begin{equation}  
    \left\{\begin{matrix}L(\theta^g)=\mathbb{E}[(y-q_i^{t})^2],\\ y=r_i^{t}+\gamma q_i'^{(t+1)},\end{matrix}\right.\label{eq19}
\end{equation}
 where $ r_i^{t}$ is the reward obtained by the robot $i$ at time $t$, and $\gamma$ is the discount factor. 

The shared policy network $\bar{\mu}$ of robot $i\in\mathcal{N}$  is updated through gradient ascent, aiming to maximize the agent's expected return. The gradient  update is performed according to the following equation:
\begin{equation} 
    \nabla_{\theta^a}J(\theta^a)=\mathbb{E}[\nabla_{\theta^a}\bar{\mu}(a_i|o_i)\nabla_{a_i}q_i^{t}].\label{eq20}
\end{equation} 
It is crucial to note that, due to the utilization of a shared network approach, in the update process, each robot is required to update the network parameters sequentially based on the sampled data. The subsequent section will present a comprehensive explanation of the specific update process. 

\subsection{\blue{Overview of the LIA MADDPG Framework} }\label{subsection b}

\blue{The LIA\_MADDPG framework consists of two distinct phases: the off-line centralized training phase and the on-line distributed execution phase. To enhance the learning process, the framework employs four types of neural networks: an actor network for generating the action policy; a critic network for evaluating the action policy; a target actor network; and a target critic network for stabilizing the learning process. Additionally, a Local Information Aggregation (LIA) module is integrated to accelerate the learning process. The overall structure of the LIA\_MADDPG algorithm is illustrated in Fig. 3, and the pseudocode is summarized in Algorithm \ref{alg1}. In the following sections, we will provide a detailed explanation of these two phases.}

 \begin{figure*}[h]
\centering
\includegraphics[width=0.95\textwidth]{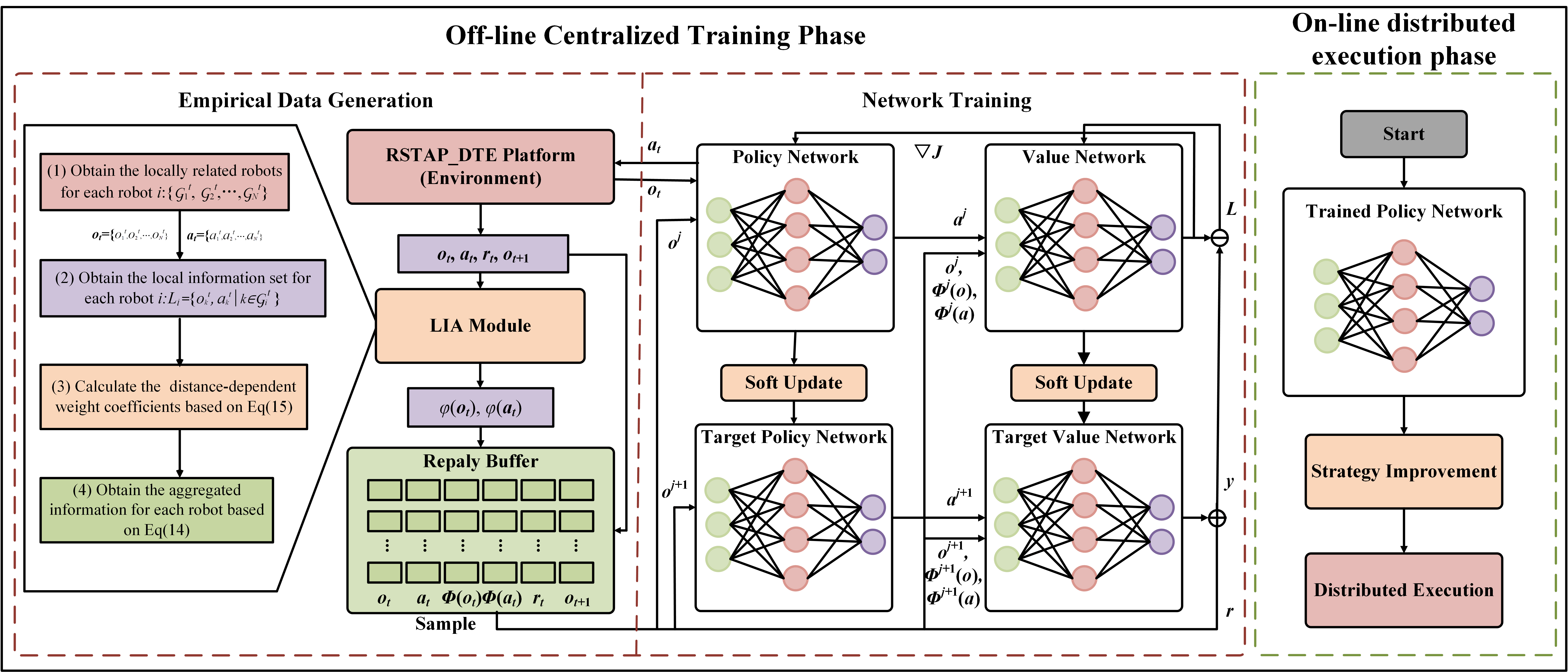}
\caption{\blue{Structure of the proposed LIA\_MADDPG, which consists of two parts:  Off-line centralized training and on-line distributed execution. The LIA\_MADDPG employs four types of neural networks: 1) an actor network for generating the action policy; 2) a critic network for evaluating the
action policy; 3) a target actor network; 4) a target critic network for stabilizing the learning process.
Furthermore, a local information aggregation (LIA) module is employed to accelerate the learning process.}}
\label{main_network}
\end{figure*}

\paragraph{\textbf{\blue{Off-line} Centralized Training Phase}}\label{subsection b.a}

The \blue{off-line} centralized training process can be divided into the generation of empirical data and network training, which are executed alternately. 

\textbf{Empirical Data Generation:} \blue{As depicted in the leftmost part of Fig. 3, the empirical data generation phase involves each robot continuously interacting with the environment to collect relevant experiential data. During this process, each robot determines its actions $a_i$ based on its current local observation $o_i$ using a shared policy network. Moreover, each robot identifies its set of related robots $\mathcal{G}_{i}^t$ and its local information set $L_i$ according to the previously described method. The LIA module then aggregates this information, applying distance-dependent weight coefficients to emphasize the influence of closer neighbors. This aggregation process yields a comprehensive dataset that encapsulates both individual and collective behavior within the environment. The collected data $(\bm{o}, \bm{a}, \bm{\phi}(\bm{o}), \bm{\phi}(\bm{a}), r, \bm{o}')$ are subsequently stored in the experience replay buffer $\mathcal{D}$ , which serves as a critical resource for optimizing the policy and value networks during the network training phase.} This process of generating empirical data can be found in lines 6-12 of Algorithm \ref{alg1}.

\textbf{Network Training:} After the experience data is generated, the data set is extracted from the buffer $\mathcal{D}$ for training based on priority experience replay \cite{schaul2015prioritized}. At this stage, we need to update $\bar{G}$ and $\bar{\mu}$ according to \eqref{eq19} and \eqref{eq20}. First, use the target $\bar{G}'$ to calculate the extended Q function $q^{t+1}$ of the next state, and use the temporal difference method to optimize the parameters of the $\bar{G}$ network based on the value of the extended Q function. Then, use $\bar{G}$ to calculate the extended Q value $q^{t}$ of the current state-action, and use the calculated Q value and the action $a^{t+1}$ output by $\bar{\mu}$ to update the parameters of the $\bar{\mu}$ network (Algorithm \ref{alg1} Lines 13-17).

\begin{algorithm}[h]
\SetKwInOut{Input}{Input}
\SetKwInOut{Output}{Output}
\caption{Training Process of LIA\_MADDPG}\label{alg1}
\Input{Max\_episode\_length $T_e$, batch size $\delta$, discount factor $\gamma$, soft target update rate $\eta$}
\Output{Trained actor network $\bar{\mu}^{*}$}
\KwSty{Initialize}: Expand \textit{Q} network $\bar{G}$ and policy network $\bar{\mu}$ with random weights $\theta^q$, $\theta^a$, target networks $\bar{{G}}^{\prime}$ and $\bar{\mu}'$  with weights $\theta'^g\leftarrow\theta^g$,$\theta'^a\leftarrow\theta^a$, and replay buffer $\mathcal{D}$

\While{Training is not terminated}{
    Initialize a random process $\mathcal{H}$ for action exploration\;
    Receive initial state $\bm{o} = (o_1,o_2,\dots, o_N)$\;
    \For{\textit{t} = 1 to $T_e$}{
        \textbf{for} each robot \textit{i}, \textbf{do} $ a_i=\bar{\mu}
                    (o_i;\theta^a)+\mathcal{H}_t$\;
        Execute actions $\bm{a}=(a_1,a_2,\dots,a_N)$ and observe reward \textit{r} and new state $\textbf{\textit{o}}'$\;        Obtain the locally related robot set $G=(G_1,G_2,\dots, G_N)$ of each robot\;      
        Calculate $\bm{\phi}(\bm{o})=(\varphi_1(o_k)\Big|_{k\in\mathcal{G}_1},\varphi_2(o_j)\Big|_{k\in\mathcal{G}_2},\dots,\varphi_N(o_k)\Big|_{k\in\mathcal{G}_N})$ according to \eqref{eq14}\;  
        Calculate $\bm{\phi}(\bm{a})=(\varphi_1(a_k)\Big|_{k\in\mathcal{G}_1},\varphi_2(a_k)\Big|_{k\in\mathcal{G}_2},\dots,\varphi_N(a_k)\Big|_{k\in\mathcal{G}_N})$ according to \eqref{eq15}\; 
        Store $(\bm{o}, \bm{a}, \bm{\phi}(\bm{o}), \bm{\phi}(\bm{a}), r, \bm{o}')$ in replay buffer $\mathcal{D}$\;
        $\bm{o}\leftarrow\bm{o}'$
        
        \For{\textit{I} = 1 to $N$}{
            Sample a random minibatch of $\delta$ samples $(\bm{o}^j, \bm{a}^j, \bm{\phi}(\bm{o})^j, \bm{\phi}(\bm{a})^j, r^j, \bm{o}'^j)$ from $\mathcal{D}$\;
            Update critic by minimizing the loss according to \eqref{eq19}\;
            Update actor using the sampled policy gradient according to \eqref{eq20}\;
        }
        Update the target networks by soft update manner:
        \begin{gather*}
        \theta'^g\leftarrow\eta\theta^g+(1-\eta)\theta'^g\\
        \theta'^a\leftarrow\eta\theta^a+(1-\eta)\theta'^a
        \end{gather*}
    }
}
\end{algorithm}

\paragraph{\textbf{On-line Distributed Execution Phase}}\label{subsection b.b}
LIA\_MADDPG's strength lies in its ability to autonomously develop strategies for problem-solving through extensive instance sampling and learning. This end-to-end approach enables it to quickly produce suitable task allocation solutions without complex heuristic rules. However, it struggles in dynamic environments with partial observability, as robots must adapt to variable conditions and make decisions based on incomplete information. Hence, during the distributed execution phase, strategy optimization methods are required to enhance each robot's decisions based on the shared policy network $\bar{\mu}$.
Therefore, there are two process during the distributed  execution phase: policy output (Algorithm \ref{alg2}, Lines 2-4) and policy improvement (Algorithm \ref{alg2}, Lines 5-8), which alternate to facilitate decision-making by each robot.

In the policy output stage, robot  feeds the current observation $o_i^t$   into a pretrained neural network $\bar{\mu}(\cdot|_{\theta^a})$  to obtain the corresponding action  $a_i^t$. This stage maps observations of the environment to actions with the shared policy, generating initial decisions using the pretrained network.

In the policy improvement stage, we design a deviation probability $\delta_{i,g_i^t}$  depending on two factors to assess the probability of a robot deviating from the current policy or target task. 
  Firstly, we consider the correlation between the number of robots already assigned to the current selected target task $h_{g_i^t}^t$ and the highest number of robots can be assigned to the task $\bar{h}_{g_i^t}$. If $h_{g_i^t}^t$ is closer to $\bar{h}_{g_i^t}$, there is a higher probability of deviation. This is because we encourage robots to explore tasks with fewer assigned robots.
\begin{equation}
    \bar{h}_{g_i^t}\otimes h_{g_i^t}^t=
    \begin{cases}
        \begin{aligned}
            &\bar{h}_{g_i^t}-h_{g_i^t}, && \text{if}\enspace \bar{h}_{g_i^t}>h_{g_i^t}^t\\
            &0, && \text{otherwise}\\
        \end{aligned}\label{eq26}
    \end{cases}       
\end{equation}
Secondly, the correlation between the number of neighboring robots adopting the same strategy.  Let $\alpha_i$ represent the number of neighbor agents that can be observed by robot \textit{i}, and $\beta_i$ represent the number of robots $i'$ that adopt the same strategy as robot $i$, i.e., $g_{i^{\prime}}^{t}=g_{i}^{t},i^{\prime}\in \mathcal{N}_{i}^t$.  By calculating the difference $\alpha_i-\beta_i$, we can assess the level of potential conflict with neighboring robots. In order to avoid tasks that may conflict with neighboring robots, we encourage robots to prioritize tasks where the difference between $\alpha_i$ and $\beta_i$ is smaller. This ensures that robot $i$ selects a task that is less likely to cause conflicts or overlap with the tasks of its neighboring robots. 
Finally, based on the above discussions, we design the  deviation probability  as  $\delta_{i,g_i^t}=e^{-\left(\bar{h}_{g_i^t}\otimes h_{g_i^t}^t\right)(\alpha_i-\beta_i)}$. 

\begin{algorithm}[h]
\SetKwInOut{Input}{Input}
\SetKwInOut{Output}{Output}
\caption{Distributed Policy Improvements and Execution}\label{alg2}
\Input{Robot \textit{i}'s policy network $\bar{\mu}^{*}$ , state $\bm{s_i}$ action space $\bm{a_i}$}
\Output{Optimized allocation strategy $\bar{\mu}_\text{finl}$}

\For{\textit{t} = 1 to $T_e$}{
    \tcp{Policy execution}
    \If {$c_i^t=1$}{
        obtain observation $o_i^t$ of robot \textit{i}\;
        $g_i^t\leftarrow a_i^t=\bar{\mu}^*(o_i^t|_{\theta^a})$\;
        \tcp{Policy improvement}
        Calculate divergence probability $\delta_{i,g_i^t}$\ and $\xi$=random(0,1)\;
        \If {$\xi<\delta_{i,g_i^t}$}{
            $a_i^t\leftarrow g_i^t=argmax_{j\in\mathcal{M}}\varphi_1r_{i,j}-d_{i,j}$\;
        }
        $s_i^{t+1}\leftarrow\mathcal{P}_i(s_i^t,a_i^t)$\;
    }
}
\end{algorithm}

\subsection{ Interpretability of LIA\_MADDPG}\label{subsection d}
\blue{In robotic physics systems, the interpretability of MARL is crucial for ensuring that robot teams can execute tasks safely and efficiently within physical constraints. Robots must interact not only with complex, dynamic environments but also coordinate with one another, significantly increasing the complexity of group decision-making. Therefore, understanding and explaining the mechanisms behind these decisions is vital for optimizing the system performance and ensuring that their behavior adheres to physical laws.}

\blue{Early research on the interpretability of RL primarily focused on single-agent models, employing techniques like feature importance analysis and policy-level explanations to clarify why an agent takes a particular action in a given state. These methods highlight the most critical features driving agent decisions. For example, Shapley Additive Explanations (SHAP)\cite{lundberg2017unified} have been widely used to quantify each feature’s contribution to the decision-making process. In addition, visualization techniques have also played an important role in enhancing RL model interpretability. Tools such as interactive saliency maps and visual analytics provide deep insights into agent learning strategies. For instance, \cite{wang2021visual} used saliency maps to analyze agent strategies in Atari games, while \cite{wang2018dqnviz} performed multi-level analyses on the DQN algorithm, offering detailed insights into agent behavior across different learning stages. These techniques not only help in understanding how agents learn over time but also make complex decision processes more intuitive and easier to interpret. However, MARL introduces additional complexity. In multi-agent systems, agents must learn from both environmental interactions and coordination or competition with other agents, making their behavior harder to explain. Traditional methods struggle to clarify how individual behaviors evolve into collective strategies, posing significant challenges for the design and validation of MARL systems.}

\blue{To address these challenges, this paper explores the interpretability of LIA\_MADDPG in large-scale robot task allocation, examining the algorithm's logic and reward function design. Additionally, we explore visual analytics as a promising research direction. In the last part of the experiment, we have developed a high-fidelity physics engine simulation system and employed visualization techniques to illustrate the interactions and behavioral evolution of agents under different physical conditions. These visual tools not only enhance the understanding of MARL decision-making processes but also reveal the interaction mechanisms and behavioral dynamics of agents. By tracing the decision rationale of each agent in complex physical environments, these methods provide strong theoretical and practical support for further research on interpretability in physical systems, paving the way for the development of more transparent and reliable multi-agent systems.}

\section{Experiments}\label{section4}
In this section, we will conduct a comprehensive evaluation of the proposed LIA\_MADDPG algorithm to assess its efficiency and scalability in a simulated multi-robot system. \blue{Our experiments include}: benchmarking LIA\_MADDPG against established reinforcement learning methods for dynamic task allocation and agent coordination; conducting ablation studies to highlight the contributions of the Local Information Aggregation (LIA) module and policy improvement method; \blue{integrating the LIA module with other MARL methods to showcase its wide applicability}; performing scalability tests across various robot system sizes to analyze the algorithm’s performance consistency; and utilizing a high-fidelity physics engine simulation to bridge the gap between theoretical models and practical applicability. Each component is meticulously designed to rigorously test the algorithm under diverse and controlled conditions, ensuring a robust evaluation of its potential in real-world multi-robot applications. 

\subsection{Experimental Setup}
{\bf Simulation Environment Design:} The simulation environment for the training process is designed in a 1000*1000m$^2$ obstacle-free 2D space. At the beginning of each training episode, we randomly generate the initial positions for $N$ robots and $M$ target tasks, where the coordinates of all positions are normalized to the range [0, 1]. During the generation process, the velocity of each robot is sampled from a uniform distribution [2m/s, 5m/s]. The velocity of each task is generated from a uniform distribution [0.5m/s, 1m/s], and its turning angle varies randomly within the range of [-$\pi$, $\pi$]. This means that the robots can choose their target tasks in any direction. The association distance between tasks and agents is set to $d_{bind}=30m$. The maximum demand of agents for each task, $\bar{h}_j$ is set to $\lceil N/M \rceil$, and the maximum number of neighbors with which an agent can perceive simultaneously is $\alpha_i=10$.

{\bf Training Process Configuration:} The maximum step length per episode is capped at 150 time steps. The reward parameters are defined as $\varphi_1=10$, $\varphi_2=-0.001,$ and $\varphi_3=1$, with $r_{i,j}$ values sampled uniformly between 0 and 1. For the neural networks, both the shared value network $\bar{G}$ and the shared policy network $\bar{\mu}$ consist of two hidden layers, each containing 128 neurons. Network updates are managed using the Adam optimizer with learning rates of 0.001 and 0.002, respectively. An experience replay buffer is maintained with a capacity of 5000 to facilitate efficient learning, and batch sizes for stochastic gradient descent (SGD) are set at 64, with a learning rate $\eta$ of 0.01 for the target network.

{\bf Performance Metrics:} Performance comparison among different algorithms is based on three key metrics: Normalized Average Total Utility (\textbf{\textit{NATU}}), which evaluates the effectiveness of an algorithm in task allocation by assessing overall rewards; Normalized Average Time Cost (\textbf{\textit{NATC}}), measuring task completion efficiency within a set timeframe; and Dominance Rate (\textbf{\textit{DR}}), indicating the frequency at which one algorithm outperforms others in terms of total rewards over a series of sample processes.

\subsection{Training Results and Comparisons} 

To systematically evaluate the effectiveness of our proposed LIA\_MADDPG algorithm, we conduct experiments involving different numbers of robots ($N$ = 30, 60, and 90) to test the pretraining strategies. Our objective is to assess the performance of LIA\_MADDPG in terms of efficiency and scalability, comparing it against \blue{six} existing reinforcement learning-based algorithms: Ind\_DQN \cite{mnih2013playing}, Ind\_DDDPG \cite{lillicrap2015continuous}, MAAC \cite{iqbal2019actor}, QMIX \cite{chen2021nqmix}, \blue{LINDA\_QMIX \cite{cao2023linda}}, and MAPPO \cite{yu2022surprising}. These methods are chosen due to their effectiveness in the relevant fields: Ind\_DQN and Ind\_DDDPG demonstrate strong performance in single-agent environments, while MAAC, QMIX, and MAPPO are recognized for their multi-agent coordination capabilities. \blue{Additionally, we have included the LINDA QMIX algorithm that incorporates the LINDA module, a local information decomposition mechanism. The LINDA module decomposes local information by leveraging agents' historical trajectories, enhancing team awareness and allowing agents to effectively estimate the global state even in partially observable scenarios. LINDA is particularly useful in small-scale heterogeneous environments, such as StarCraft, where agents can infer the state of entities beyond their observable range.} To ensure a fair comparison, all algorithms employed a unified network architecture and identical hyperparameters.
\begin{figure*}[h]
  \centering
  \includegraphics[width=1\textwidth]{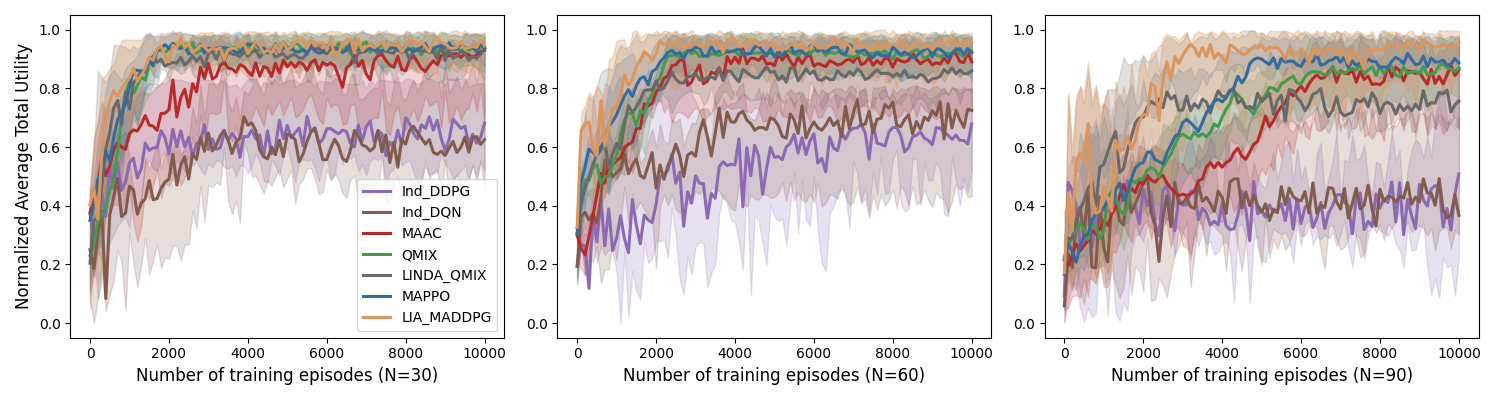}
  \caption{ \blue{Normalized average episode utility by LIA\_MADDPG and six comparison RL-based methods in different robot scales. The solid and dashed lines show the mean and standard deviation of the results over ten runs, respectively.}}
  \label{cov_fig}
\end{figure*}

The convergence curves depicted in Fig.\ref{cov_fig} show that LIA\_MADDPG consistently achieves near-optimal normalized average episode rewards across all tested robot counts, highlighting its robustness in managing the increasing complexity of state and action spaces. \blue{Though other MARL methods reach comparable performance at $N = 30$, they suffer from slower convergence, greater reward variability, and lower average rewards overall. In particular, LINDA\_QMIX, despite quickly attaining 78\% of the optimal normalized reward at $N = 90$, shows no further reward improvement during training, with significant fluctuations in its convergence curve. This is due to the large number of robots and the relatively homogenous state information in our scenario, which hampers LINDA’s ability to effectively decompose local information and develop individual teammate awareness.} On the other hand, Ind\_DQN and Ind\_DDPG, which ignore inter-agent interactions, deliver the poorest performance, underscoring the crucial role of such dynamics in ensuring effective robot coordination. Our findings reaffirm LIA\_MADDPG’s superior convergence speed, stability, and overall performance, especially in large-scale environments.

\begin{table*}[t]
\centering
\caption{\centering \scshape{\blue{Statistical comparison of the trained policies with LIA\_MADDPG (Ours), IND\_DQN, IND\_DDPG, MAAC, QMIX, LINDA\_QMIX, and MAPPO}}}
\label{tab1}
\resizebox{0.95\textwidth}{!}{%
\begin{tabular}{@{}lcccccccc@{}}
\toprule
Metrics & \textit{\textbf{N}} & \textbf{Ind DDPG} & \textbf{Ind DQN} & \textbf{MAAC} & \textbf{QMIX} & \textbf{LINDA\_QMIX} & \textbf{MAPPO} & \textbf{LIA\_MADDPG} \\ \midrule

\multirow{3}{*}{\textit{\textbf{NATU}}} 
& 30 & $0.458^{\pm 0.153}$ & $0.454^{\pm 0.185}$ & $0.742^{\pm 0.087}$ & $0.769^{\pm 0.080}$ & $\underline{\textit{0.778}}^{\pm \textit{0.081}}$ & $\textit{0.775}^{\pm \textit{0.075}}$ & $\textbf{0.793}^{\pm \textbf{0.079}}$ \\

& 60 & $0.414^{\pm 0.145}$ & $0.405^{\pm 0.187}$ & $0.770^{\pm 0.072}$ & $\textit{0.806}^{\pm \textit{0.067}}$ & $0.801^{\pm 0.064}$ & $\underline{\textit{0.828}}^{\pm \textit{0.057}}$ & $\textbf{0.864}^{\pm \textbf{0.059}}$ \\

& 90 & $0.282^{\pm 0.127}$ & $0.295^{\pm 0.145}$ & $0.666^{\pm 0.079}$ & $\textit{0.730}^{\pm \textit{0.073}}$ & $0.705^{\pm 0.062}$ & $\underline{\textit{0.775}}^{\pm \textit{0.060}}$ & $\textbf{0.863}^{\pm \textbf{0.052}}$ \\

\multirow{3}{*}{\textit{\textbf{NATC}}} 
& 30 & $0.395^{\pm 0.179}$ & $0.387^{\pm 0.187}$ & $0.372^{\pm 0.187}$ & $0.357^{\pm 0.187}$ & $\underline{\textit{0.331}}^{\pm \textit{0.188}}$ & $\textit{0.348}^{\pm \textit{0.185}}$ & $\textbf{0.325}^{\pm \textbf{0.123}}$ \\

& 60 & $0.529^{\pm 0.182}$ & $0.527^{\pm 0.187}$ & $0.460^{\pm 0.173}$ & $\textit{0.431}^{\pm \textit{0.172}}$ & $0.451^{\pm 0.171}$ & $\underline{\textit{0.418}}^{\pm \textit{0.170}}$ & $\textbf{0.364}^{\pm \textbf{0.154}}$ \\

& 90 & $0.447^{\pm 0.171}$ & $0.450^{\pm 0.174}$ & $0.397^{\pm 0.165}$ & $\textit{0.362}^{\pm \textit{0.160}}$ & $0.378^{\pm 0.164}$ & $\underline{\textit{0.309}}^{\pm \textit{0.142}}$ & $\textbf{0.260}^{\pm \textbf{0.138}}$ \\

\multirow{3}{*}{\textit{\textbf{DR}}} 
& 30 & 2\% & 3\% & 12\% & 10\% & \textit{15\%} & \underline{\textit{25\%}} & \textbf{33\%} \\

& 60 & 1\% & 2\% & 6\% & \textit{14\%} & 9\% & \underline{\textit{23\%}} & \textbf{45\%} \\

& 90 & 0\% & 0\% & 2\% & \textit{7\%} & 3\% & \underline{\textit{10\%}} & \textbf{78\%} \\ \midrule

\end{tabular}%
}
\end{table*}

{\bf Comparison with existing methods:} Following the training phase, we carry out an extensive evaluation of our policy's performance through 100 distinct initial test scenarios, characterized by randomly assigned task locations and robot departure points. These tests span three different robot populations sizes (N=30, 60, and 90). As detailed in Table \ref{tab1}, our method consistently outperforms traditional MARL methods, especially in the Normalized Average Total Utility (NATU) metric. Though the other methods are competitive at a smaller scale (N=30), our approach demonstrates significant superiority at larger scales, achieving NATU values up to twice as high as those of individualistic algorithms like Ind\_DQN and Ind\_DDPG at N=90.

Regarding the Normalized Average Time Cost (NATC), our method demonstrates the lowest time consumption across all scales, underscoring its superior efficiency compared to both single-agent and other multi-agent strategies. This emphasizes the robustness and consistent performance of our algorithm. In terms of the Dominance Ratio (\textbf{\textit{DR}}), our method slightly outperforms other multi-agent reinforcement learning algorithms at smaller scales and shows a significant advantage as the scale increases. At $N = 90$, our method achieves a \textbf{\textit{DR}} of nearly 80\%, demonstrating its effectiveness and scalability. In stark contrast, methods like MAPPO experience a notable decline in performance as $N$ increases, with \textbf{\textit{DR}} values dropping significantly—MAPPO’s \textbf{\textit{DR}} falls to just 10\% at $N = 90$. \blue{LINDA-enhanced QMIX (LINDA\_QMIX) faces similar issues, and in fact, the problem is even more pronounced. This further supports the idea that the LINDA module is particularly useful in small-scale heterogeneous environments like StarCraft. However, in our problem setting, agents only need to focus on nearby relevant agents when selecting target tasks, making global state awareness unnecessary.} The experimental results demonstrate that our method not only maintains high efficiency in task completion across varying scales but also significantly improves performance metrics compared to existing algorithms, affirming its superiority across diverse operational contexts.

\subsection{Performance Analysis of Key Components}
In this subsection, we investigate the effects of the Local Information Aggregation (LIA) module and the policy improvement techniques as detailed in Section \ref{section3}. To assess their contributions, we conducted two sets of ablation experiments. 

\begin{table*}[t]
\centering
\caption{\centering\scshape{\blue{Statistical comparison of the policies learned with three enhanced MARL algorithms (LIA\_MADDPG, LIA\_MAPPO, and LIA\_MAAC) and there traditional visions (MADDPG, MAPPO, and MAAC)}}}
\label{tab_lia}
\resizebox{0.85\textwidth}{!}{%
\begin{tabular}{@{}lc|cc|cc|cc@{}}
\toprule
Metrics & \textit{\textbf{N}} & MAAC & LIA\_MAAC & MAPPO & LIA\_MAPPO & MADDPG & \begin{tabular}[c]{@{}c@{}}LIA\_MADDPG\\ \end{tabular} \\ \midrule

\multirow{3}{*}{\textit{\textbf{NATU}}} 

& 30 
& $0.742^{\pm 0.086}$ & $\textbf{0.784}^{\pm \textbf{0.117}}$ 
& $0.774^{\pm 0.075}$ & $\textbf{0.803}^{\pm \textbf{0.067}}$ 
& $0.749^{\pm 0.108}$ & $\textbf{0.793}^{\pm \textbf{0.078}}$ \\

& 60 
& $0.770^{\pm 0.072}$ & $\textbf{0.835}^{\pm \textbf{0.062}}$ 
& $0.828^{\pm 0.057}$ & $\textbf{0.859}^{\pm \textbf{0.063}}$  
& $0.782^{\pm 0.105}$ & $\textbf{0.864}^{\pm \textbf{0.059}}$ \\

& 90 
& $0.666^{\pm 0.079}$ & $\textbf{0.755}^{\pm \textbf{0.065}}$ 
& $0.774^{\pm 0.060}$ & $\textbf{0.884}^{\pm \textbf{0.065}}$ 
& $0.693^{\pm 0.141}$ & $\textbf{0.863}^{\pm \textbf{0.052}}$ \\

\multirow{3}{*}{\textit{\textbf{NATC}}} 
& 30 
& $0.372^{\pm 0.187}$ & $\textbf{0.332}^{\pm \textbf{0.136}}$ 
& $0.348^{\pm 0.185}$ & $\textbf{0.303}^{\pm \textbf{0.112}}$  
& $0.356^{\pm 0.127}$ & $\textbf{0.325}^{\pm \textbf{0.123}}$ \\

 & 60
 & $0.460^{\pm 0.173}$ & $\textbf{0.382}^{\pm \textbf{0.127}}$ 
 & $0.418^{\pm 0.170}$ & $\textbf{0.336}^{\pm \textbf{0.112}}$  
 & $0.464^{\pm 0.166}$ & $\textbf{0.364}^{\pm \textbf{0.154}}$ \\
 
& 90 
& $0.397^{\pm 0.165}$ & $\textbf{0.296}^{\pm \textbf{0.124}}$ 
& $0.309^{\pm 0.142}$ & $\textbf{0.273}^{\pm \textbf{0.110}}$  
& $0.356^{\pm 0.161}$ & $\textbf{0.260}^{\pm \textbf{0.138}}$ \\

\multirow{3}{*}{\textit{\textbf{DR}}} 
 & 30 & 42\% & \textbf{58\%} & 38\% & \textbf{62\%}  & 35\% & \textbf{65\%} \\
 & 60 & 31\% & \textbf{69\%} & 26\% & \textbf{74\%}  & 20\% & \textbf{80\%} \\
 & 90 & 25\% & \textbf{75\%} & 12\% & \textbf{88\%}  & 10\% & \textbf{90\%} \\ \midrule
\end{tabular}%
}
\end{table*}

\begin{figure}[h]
\centering
\includegraphics[width=0.48\textwidth]{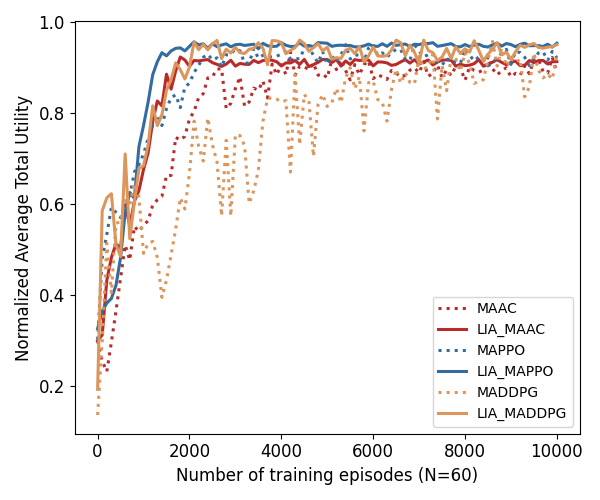}
\caption{\centering\scshape{\blue{Normalized average episode utility by three enhanced MARL algorithms (LIA\_MADDPG, LIA\_MAPPO, and LIA\_MAAC) and their traditional visions (MADDPG, MAPPO, and MAAC) in 60 robot scales.}}}
\label{lia_fig}
\end{figure}

\blue{We first compare the performance of LIA\_MADDPG with the traditional MADDPG algorithm, which does not include the LIA module. Additionally, we integrate the LIA module into two other MARL algorithms, MAAC and MAPPO, to evaluate its impact across different frameworks. The results, illustrated in Fig.\ref{lia_fig} and Table \ref{tab_lia}, clearly demonstrate the superior performance of the LIA-enhanced algorithms across various metrics. Specifically, the LIA module significantly accelerates the convergence process of the base algorithms, as evidenced by the rapid increase in normalized average total utility (NATU) for LIA\_MADDPG, LIA\_MAPPO, and LIA\_MAAC during the initial training phases, as shown in Fig.\ref{lia_fig}. Moreover, the LIA module improves training stability, as reflected in the reduced fluctuations in the utility curves of the LIA-enhanced algorithms compared to their traditional counterparts. We have also conducted a statistical evaluation of these algorithm pairs using 100 sets of randomly generated test cases. The statistical results in Table \ref{tab_lia} further corroborate these findings. The LIA-enhanced algorithms consistently outperform their traditional versions across all evaluated metrics, including NATU, NATC, and DR. Notably, as the problem scale increases, the advantages of the LIA module become even more pronounced. These results underscore the effectiveness of the LIA module in optimizing multi-agent reinforcement learning algorithms, suggesting its broader applicability in complex, dynamic environments.}

\begin{figure}
    \centering
    \begin{subfigure}{0.24\textwidth}
        \centering
        \includegraphics[width=\linewidth]{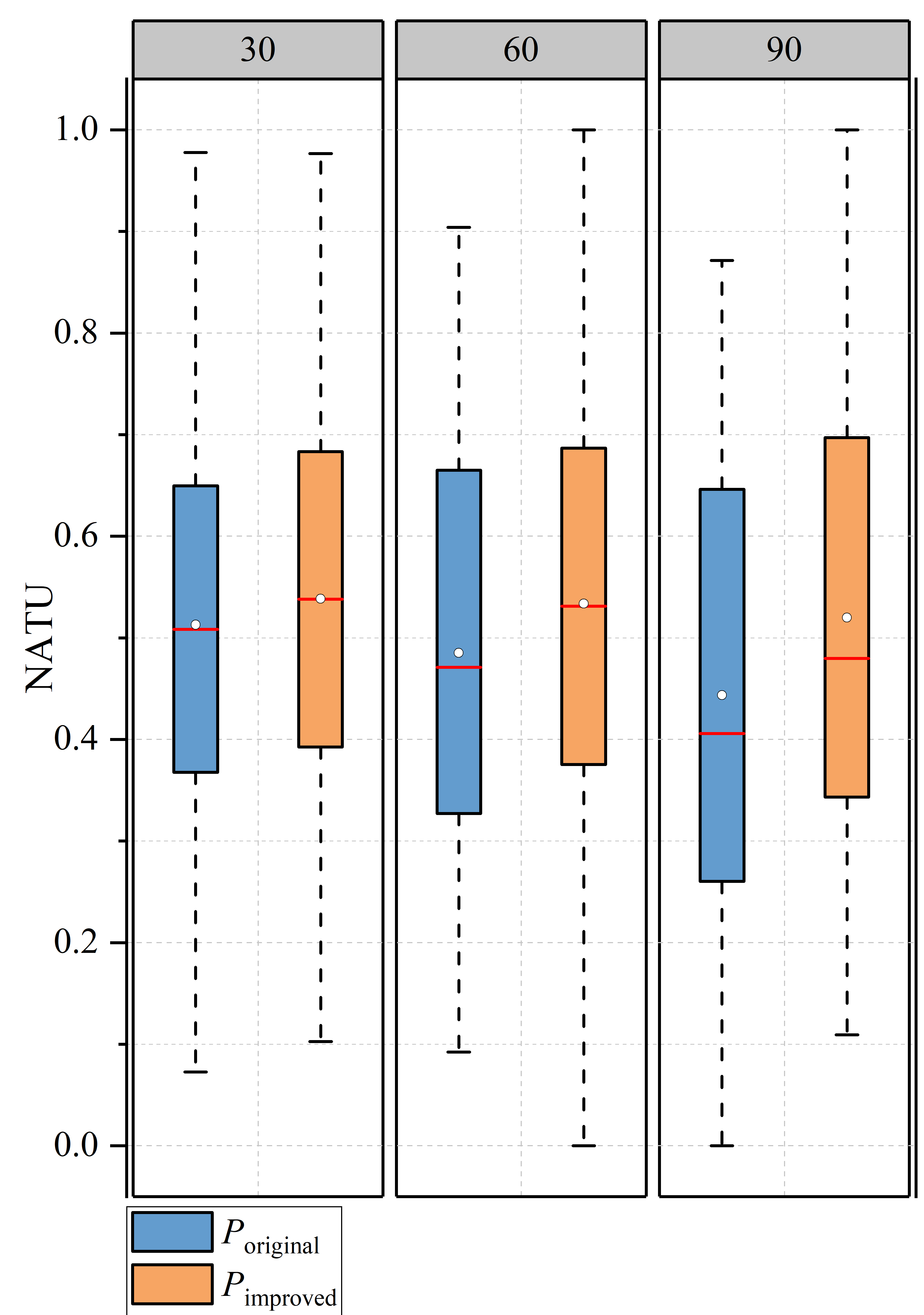}
            \caption{}
        \label{C_NATU}
    \end{subfigure}%
    \begin{subfigure}{0.24\textwidth}
        \centering
        \includegraphics[width=\linewidth]{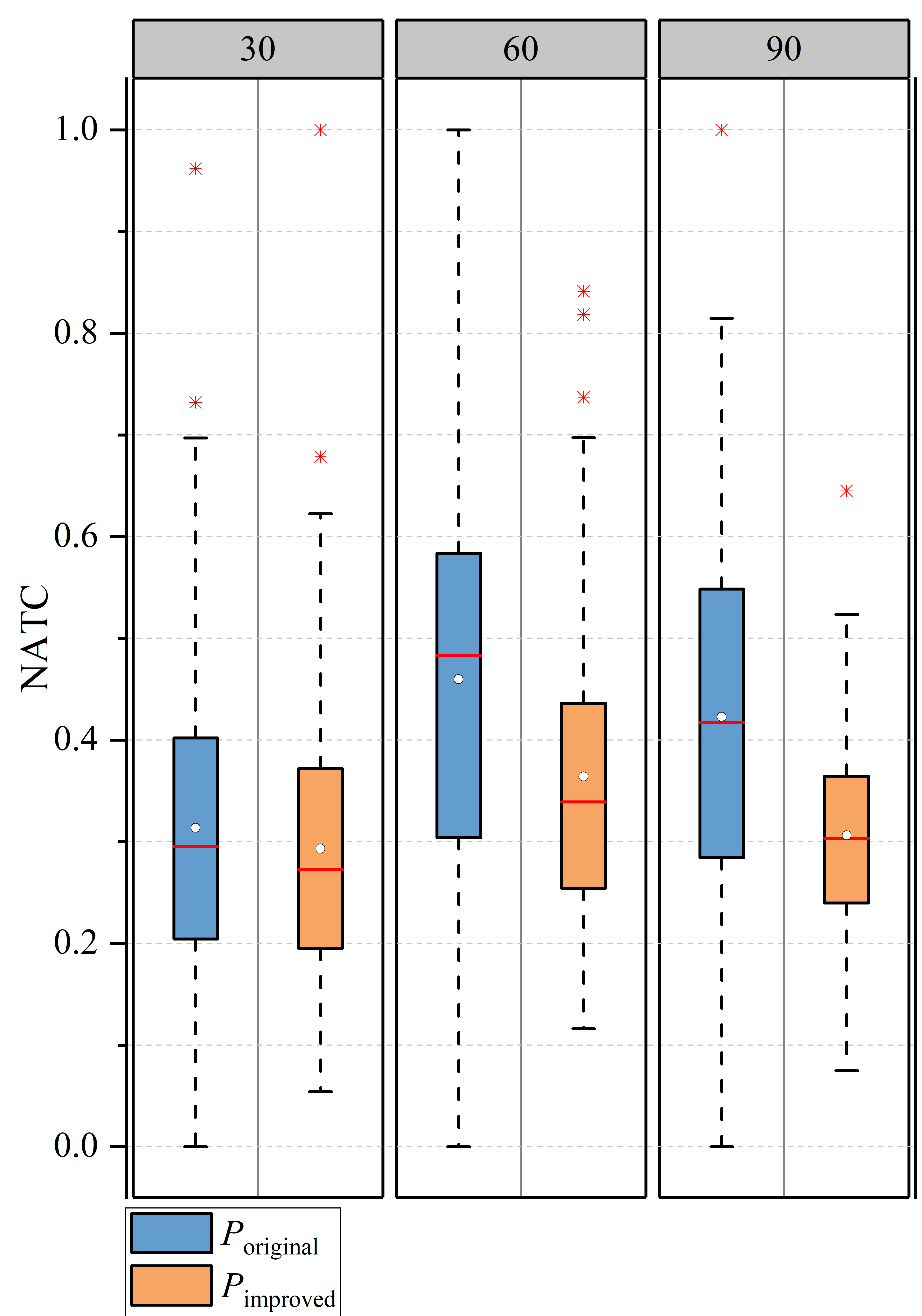}
        \caption{}
        \label{C_NATC}
    \end{subfigure}
    \caption{\blue{The statistics of \textbf{\textit{NATU}} and \textbf{\textit{NATC}} between our LIA\_MADDPG and LIA\_MADDPG without policy improvement across different numbers of robots.}}
    \label{c_fig}
\end{figure}

Subsequently, \blue{to verify the effectiveness of the policy improvement method during the online distributed execution phase, we conduct a comparative experiment. We compare the performance of a pretraining policy with LIA\_MADDPG, named $P_{original}$, against the same policy after incorporating the policy improvement method, named $P_{improved}$.} This comparison is implemented through testing both algorithms across 100 randomly generated scenarios, with the outcomes documented in Fig.\ref{c_fig} and Table \ref{tab_nLIA}. The empirical evidence shows that \blue{$P_{improved}$} attains higher Normalized Average Task Utility (\textit{\textbf{NATU}}) and reduced Normalized Average Task Completion Time (\textit{\textbf{NATC}}) compared to \blue{$P_{original}$}. This difference highlights the significant role of the policy improvement method in encouraging robots to actively explore and adjust their strategies to optimize utility. Additionally, the task completion rate (\textit{\textbf{DR}}), detailed in the third column of Table \ref{tab_nLIA}, indicates that the version lacking the strategy improvement method performs adequately only in smaller settings ($N=30$). However, as the number of robots increases, its performance notably deteriorates, underscoring the policy improvement method's essential role in boosting algorithm performance under more complex and challenging conditions. These findings robustly demonstrate the critical importance of the policy improvement method, particularly in effectively scaling up the system to handle larger scenarios. This underlines that integrating such methods is crucial for enhancing the adaptability and efficiency of the system in dynamic and unpredictable environments.

\begin{table}[t]
\centering
\caption{\centering\scshape{\blue{Comparison between LIA\_MADDPG and LIA\_MADDPG without policy improvement}}}
\label{tab_nLIA}
\resizebox{0.85\columnwidth}{!}{%
\begin{tabular}{@{}lccc@{}}
\toprule
Metrics & $N$ & $P_{original}$ & \begin{tabular}[c]{@{}c@{}}$P_{improved}$\\ \end{tabular} \\ \midrule

\multirow{3}{*}{\textit{\textbf{NATU}}}
& 30 & $0.513^{\pm 0.207 }$ & $\textbf{0.538}^{\pm \textbf{0.198}}$ \\

& 60  & $0.484^{\pm 0.064}$ & $\textbf{0.534}^{\pm \textbf{0.219}}$ \\
 
& 90 & $0.443^{\pm 0.045}$ & $\textbf{0.519}^{\pm \textbf{0.226}}$ \\

\multirow{3}{*}{\textit{\textbf{NATC}}} 
& 30 & $0.314^{\pm 0.168}$ & $\textbf{0.293}^{\pm \textbf{0.136}}$ \\

 & 60 & $0.4591^{\pm 0.154}$ & $\textbf{0.364}^{\pm \textbf{0.146}}$  \\
 
& 90 & $0.422^{\pm 0.199}$ & $\textbf{0.306}^{\pm \textbf{0.189}}$ \\

\multirow{3}{*}{\textit{\textbf{DR}}} 
 & 30 & 31\%  & \textbf{69\%} \\
 & 60 & 21\% & \textbf{79\%} \\
 & 90 &  8\% & \textbf{92\%} \\ \midrule
\end{tabular}
}
\end{table}

\subsection{Scalability Performance of LIA\_MADDPG}
In order to rigorously test the scalability of our LIA\_MADDPG method across various sizes of robotic systems, we employ a structured experimental approach. We initially train our model on a medium-scale system with \textit{N}=60 robots and \textit{M}=10 tasks. Subsequently, we test this trained model on three different scales: a small-scale system (\textit{N}=30, \textit{M}=5), a medium-scale system (\textit{N}=100, \textit{M}=8), and a large-scale system (\textit{N}=300, \textit{M}=10). Each of these scales is subjected to 100 varied initial test scenarios to ensure comprehensive comparative analysis.

For our baseline comparison, we selected the well-established greedy choice method \cite{vince2002framework}. This heuristic method processes decision-making at an individual robot level, aiming to maximize personal rewards based on real-time task status, without considering the interaction and cooperation among robots. In contrast, our LIA\_MADDPG method integrates these interactions, which is crucial for coordinated tasks in multi-robot systems. To ensure a fair evaluation, both methods are implemented with the same utility function and identical initial conditions for each test case.

\begin{figure}[h]
\centering
\includegraphics[width=0.49\textwidth]{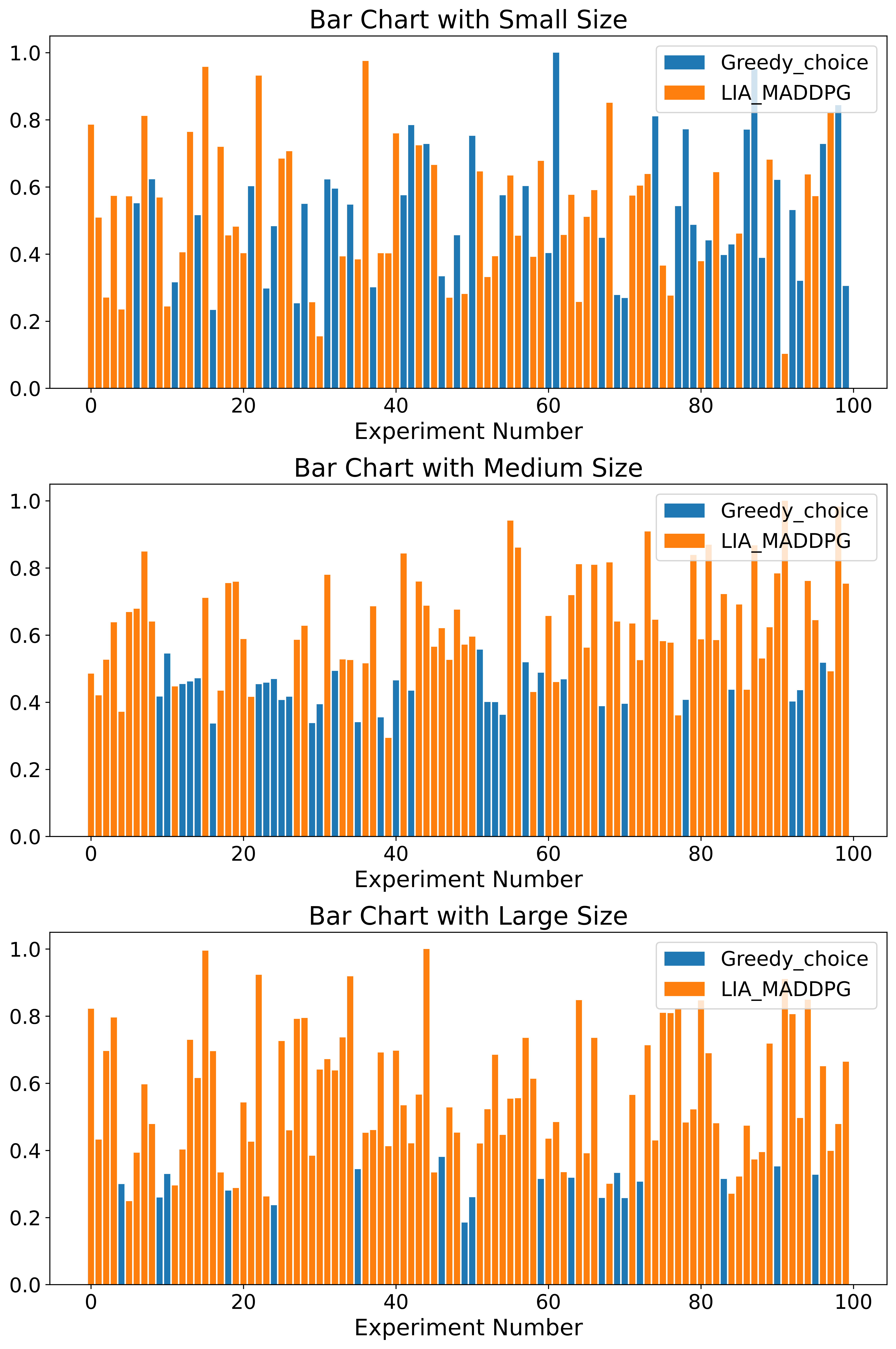}
\caption{\blue{ The bar plot shows the \textbf{\textit{DR}}   statistics comparing our method (LIA\_MADDPG) with the greedy choice method across different scales (small, medium, large). Each subfigure
reflects performance over 100 experiments, where the bar color
indicates the better-performing algorithm }}
\label{bar_fig}
\end{figure}

The effectiveness and performance metrics, specifically the task completion rate (\textit{\textbf{DR}}), normalized average task utility (\textit{\textbf{NATU}}), and normalized average task completion time (\textit{\textbf{NATC}}), are carefully analyzed. The results are presented in Table \ref{tab3}, and the \textit{\textbf{DR}} results are displayed in a bar chart in Fig. \ref{bar_fig}. 

\begin{table}[t]
\centering
\caption{\centering\scshape{\blue{Comparison between LIA\_MADDPG and the greedy choice method in different scale scenarios}}}
\label{tab3}
\resizebox{0.85\columnwidth}{!}{%
\begin{tabular}{@{}lccc@{}}
\toprule
Metrics & Scale & Greedy choice & \begin{tabular}[c]{@{}c@{}}LIA\_MADDPG\\ \end{tabular} \\ \midrule

\multirow{3}{*}{\textit{\textbf{NATU}}} & Small & $0.498^{\pm 0.207 }$ & $\textbf{0.525}^{\pm \textbf{0.198}}$ \\

& Medium  & $0.420^{\pm 0.064}$ & $\textbf{0.534}^{\pm \textbf{0.219}}$ \\
 
& Large & $0.310^{\pm 0.045}$ & $\textbf{0.509}^{\pm \textbf{0.226}}$ \\

\multirow{3}{*}{\textit{\textbf{NATC}}} & Small & $0.338^{\pm 0.168}$ & $\textbf{0.283}^{\pm \textbf{0.136}}$ \\

 & Medium & $0.331^{\pm 0.154}$ & $\textbf{0.238}^{\pm \textbf{0.146}}$  \\
 
& Large & $0.461^{\pm 0.199}$ & $\textbf{0.336}^{\pm \textbf{0.189}}$ \\

\multirow{3}{*}{\textit{\textbf{DR}}} & Small & 42\%  & \textbf{58\%} \\
 & Medium  & 32\% & \textbf{68\%} \\
 & Large & 18\% & \textbf{72\%} \\ \midrule
\end{tabular}
}
\end{table}

Regarding \textit{\textbf{NATU}} and \textit{\textbf{NATC}},  it can be seen from Table \ref{tab3} that our method consistently outperforms the greedy choice method across all scale scenarios. Notably, in the small and medium scale scenarios, our method exhibits nearly a 15\% improvement in \textit{\textbf{NATU}} and a 20\% reduction in \textit{\textbf{NATC}} compared to the greedy choice method.  Even more, in the large-scale scenario, our method showcases remarkable performance, achieving a remarkable 64.2\% improvement in \textit{\textbf{NATU}} and a substantial 36.9\% reduction in \textit{\textbf{NATC}} when compared to the greedy choice method. These results underscore the exceptional scalability of our proposed method, particularly in large-scale scenarios. From the perspective of \textit{\textbf{DR}}, our method outperforms the greedy choice method in 51 out of 100 training sets in the small-scale scenario. As the number of robot increases, our method's \textit{\textbf{DR}} gradually expands over the greedy choice method.  In large-scale problems, our method even achieves an impressive \textbf{\textit{DR}} as high as 71\%, further emphasizing the substantial advantages of our approach in addressing large-scale problems.

\subsection {Physics Engine\-based Simulation Experiment }
To rigorously evaluate our algorithm's practical applicability in real robot swarms, we have implemented a high-fidelity physics engine that features an accurate system model. This model is crucial for facilitating the 'sim-to-real' transition, allowing our algorithm to move smoothly from simulation environments to real-world applications. Our aim with these experiments is to thoroughly assess the feasibility and efficacy of the algorithm under realistic conditions.

\begin{figure}
  \centering
  \begin{subfigure}{0.49\textwidth}
    \centering
    \includegraphics[width=\linewidth]{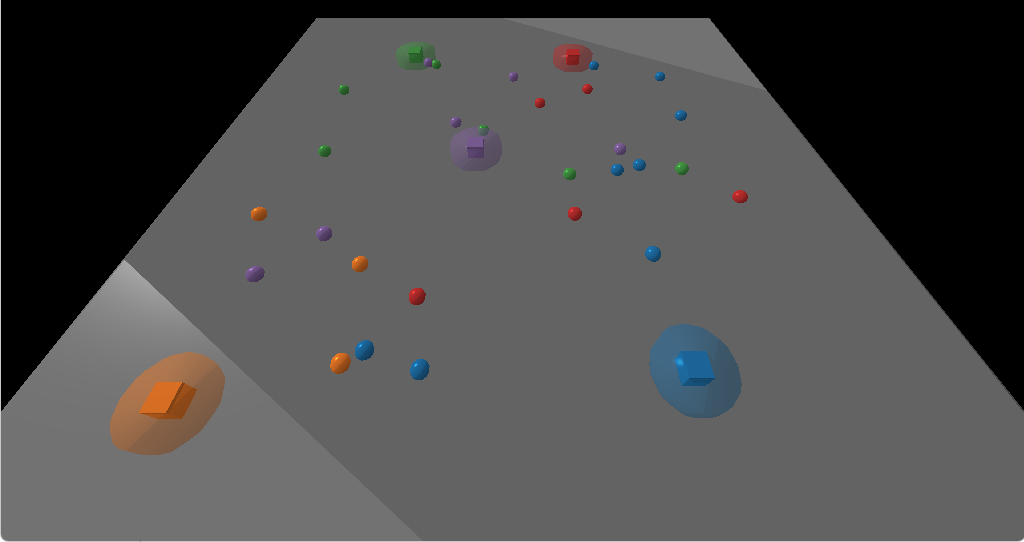}
    \caption{Small-scale scenario}
    \label{dis_fig1}
  \end{subfigure}%
  
  \begin{subfigure}{0.49\textwidth}
    \centering
    \includegraphics[width=\linewidth]{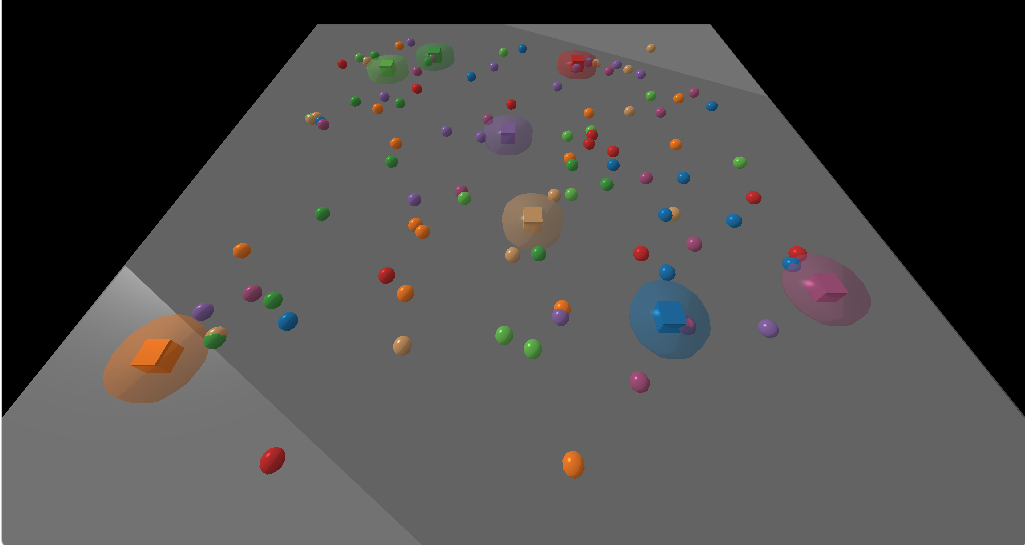}
    \caption{Medium-scale scenario}
    \label{dis_fig2}
  \end{subfigure}%
  \caption{Map of  the robot swarm task allocation problem  in a 1000 × 1000 area with small and medium scales.}
  \label{dis_fig}
\end{figure}

The simulation of the robot swarm task allocation problem is carried out using the PyBullet physics engine. The dynamics of the environment are depicted in Fig.\ref{dis_fig}, where each task is visually represented by a cube, and the state of these cubes updates dynamically. The spatial relationship between robots and tasks is indicated by transparent circles that define the association distance. Our simulation environment supports various task motion patterns, including horizontal, vertical, circular, and random movements, which contribute to the complexity of the scenario. Robots are modeled as collision-free spheres, also with dynamic state updates, enhancing the realism of their interactions.

In our subsequent experiments, we compare the performance of our algorithm with that of the greedy choice method \cite{vince2002framework}, which serves as a baseline. This comparative analysis is designed to validate our algorithm's performance in simulated scenarios that closely mimic real-world conditions. To aid in visual tracking and understanding of the task allocation process, robots change color to match their selected tasks, and this color shifts whenever they switch targets. This visual mechanism not only makes the simulation more intuitive but also provides clear, real-time feedback on the algorithm's decision-making process.

\begin{figure}
    \centering
    \begin{subfigure}{0.245\textwidth}
        \centering
        \captionsetup{justification=centering}
        \includegraphics[width=\linewidth]{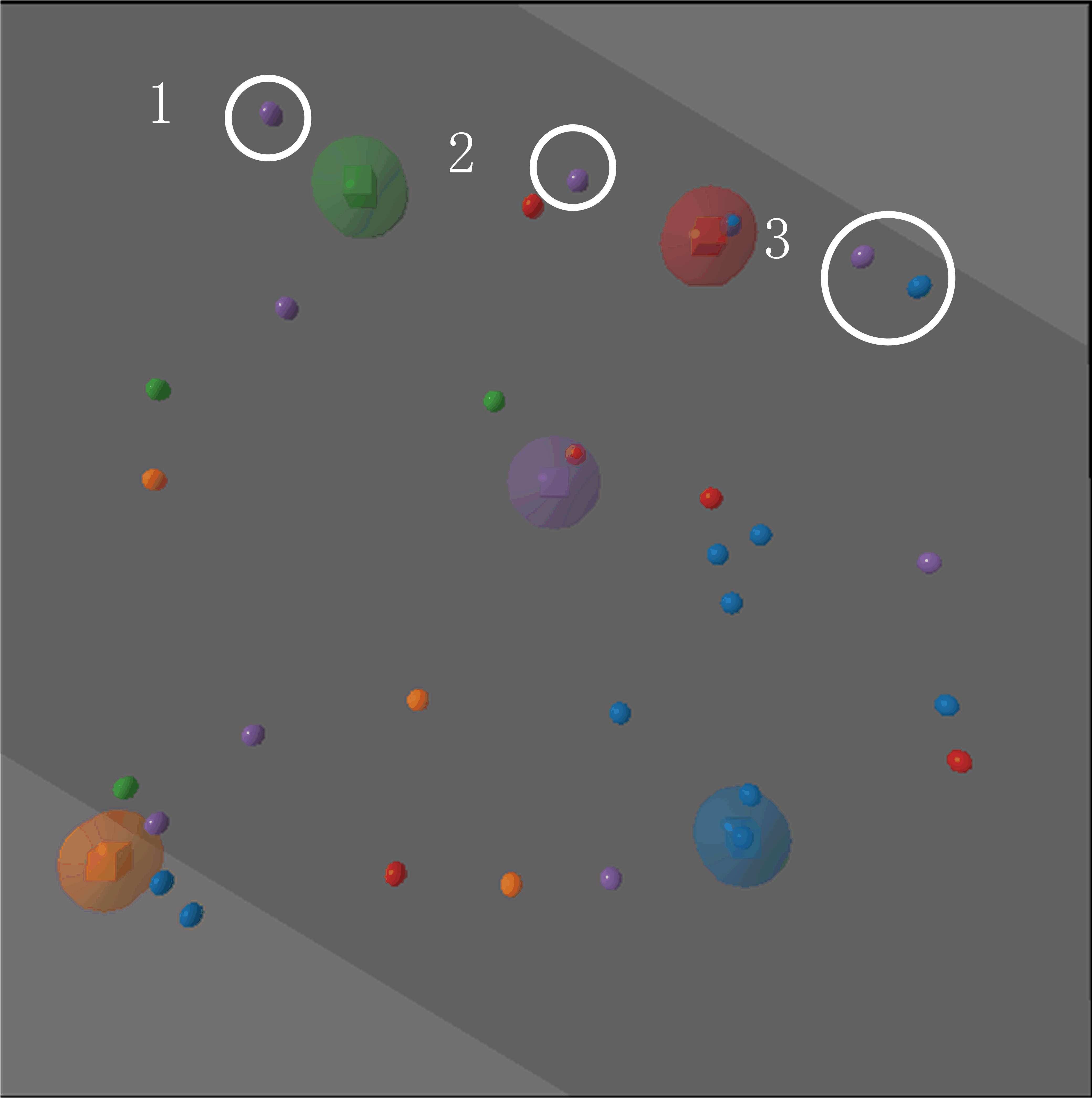}
            \caption{The greedy choice method in small-scale}
        \label{ini_fig1}
    \end{subfigure}%
    \begin{subfigure}{0.2475\textwidth}
        \centering
        \captionsetup{justification=centering}
        \includegraphics[width=\linewidth]{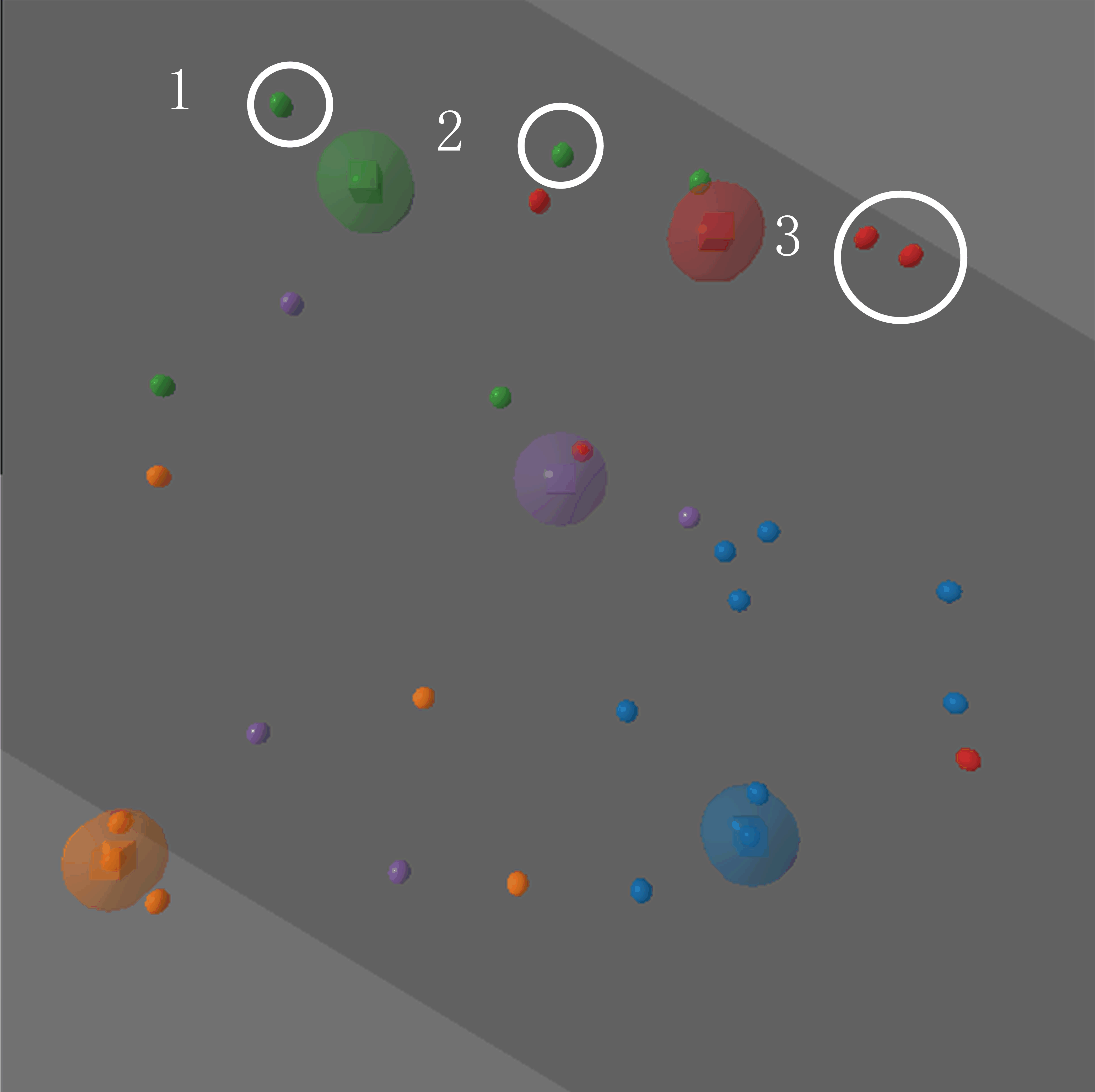}
        \caption{The LIA\_MADDPG method (ours) in small-scale}
        \label{ini_fig2}
    \end{subfigure}\\
    \begin{subfigure}{0.248\textwidth}
        \centering
        \captionsetup{justification=centering}
        \includegraphics[width=\linewidth]{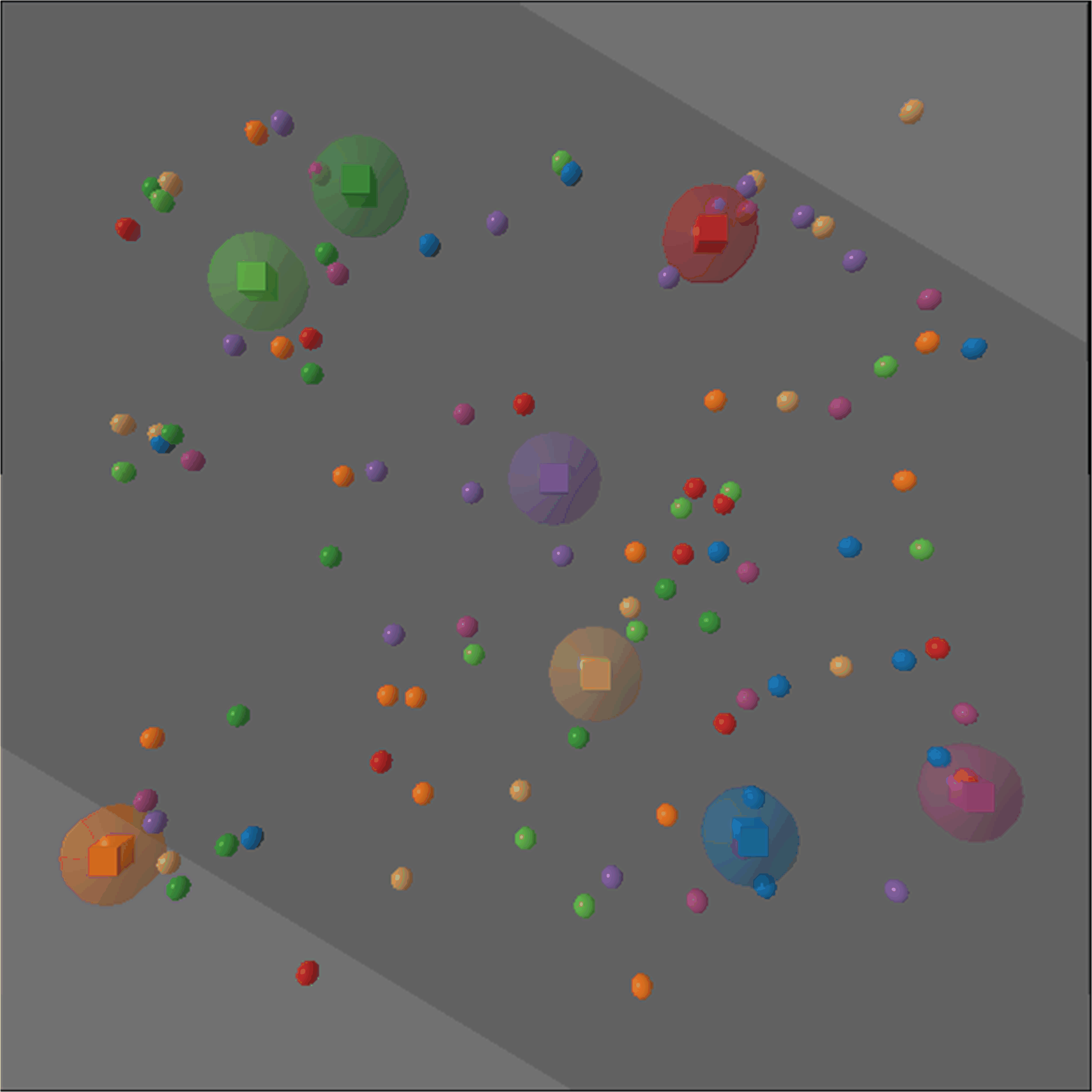}
        \caption{The greedy choice method in medium-scale}
        \label{ini_fig3}
    \end{subfigure}%
    \begin{subfigure}{0.248\textwidth}
        \centering
        \captionsetup{justification=centering}
        \includegraphics[width=\linewidth]{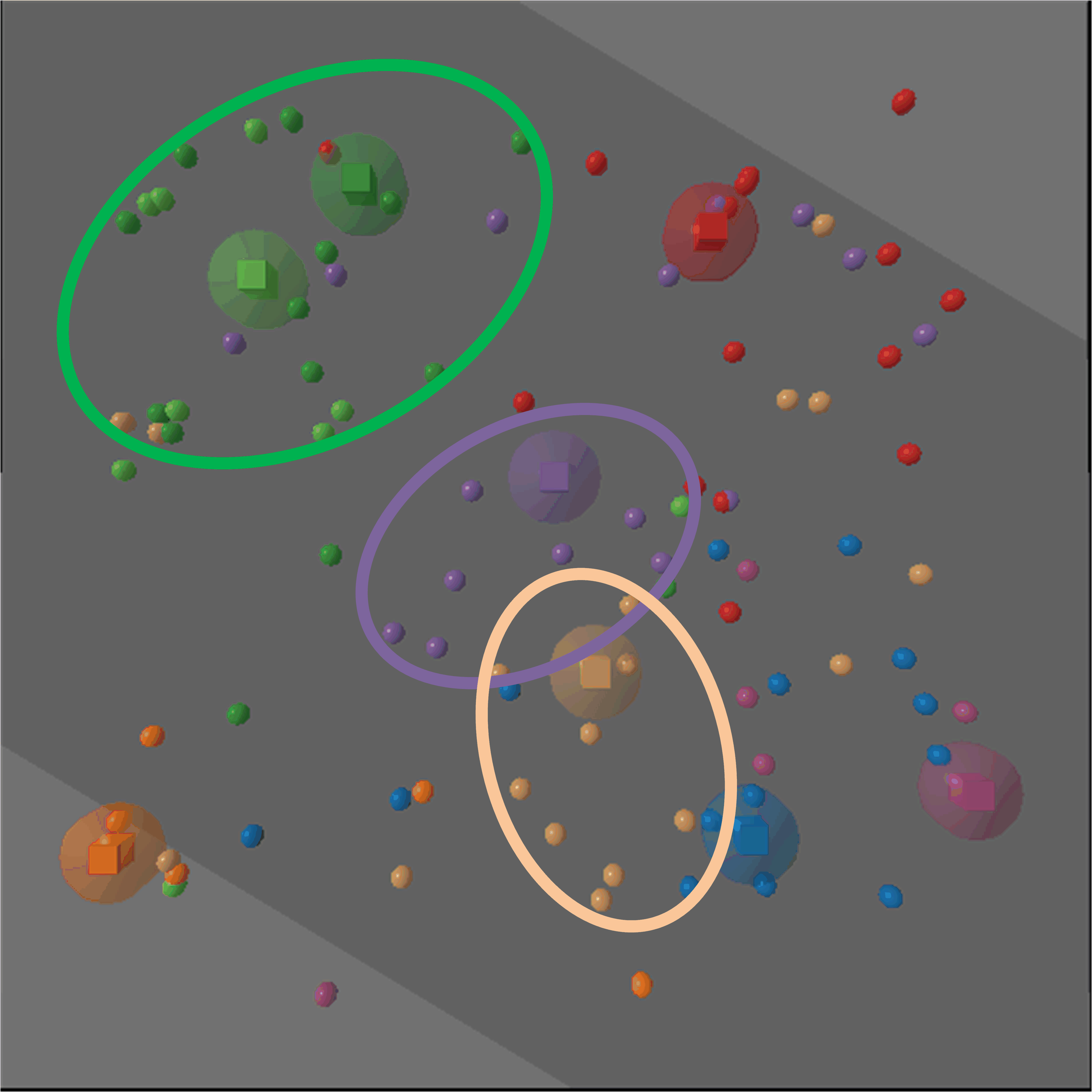}
        \caption{The LIA\_MADDPG method (ours) in medium-scale}
        \label{ini_fig4}
    \end{subfigure}
    \caption{A schematic diagram comparing the initial allocation strategies of our method and the greedy choice method in small and medium-scale scenarios.}
    \label{ini_fig}
\end{figure}

\paragraph{\textbf{Local Situational Planning}} Fig.\ref{ini_fig} provides insights into the initial allocation strategies of both algorithms in small-scale and medium-scale scenarios. Upon close examination of the figure, a notable distinction emerges: the strategy employed by the greedy method prioritizes the maximization of task rewards but tends to overlook the actual environmental conditions. For instance, as depicted in Fig.\ref{ini_fig1},   Robot 1 opts for the purple target task due to the potential    higher return,   while  it neglects the task's remote location that might entail  a risk of surpassing the agent's capacity limit. Hence, it may result in  no rewards  for Robot 1 while incurring significant costs. 
In contrast,  Fig.\ref{ini_fig2} and Fig.\ref{ini_fig4} reveal  that the initial allocation facilitated by LIA\_MADDPG leads to a more clustered arrangement of robots around their respective target task locations. This allocation strategy effectively mitigates potential risks associated with prolonged dynamic processes. Consequently, when juxtaposed with local strategies reliant on greedy selection methods, LIA\_MADDPG demonstrates a heightened ability to consider the specific context within which agents operate. It excels in adapting and making flexible decisions at the local level, effectively striking a balance between long-term returns and immediate rewards.

\begin{figure}
    \centering
    \begin{subfigure}{0.244\textwidth}
        \centering
        \includegraphics[width=\linewidth]{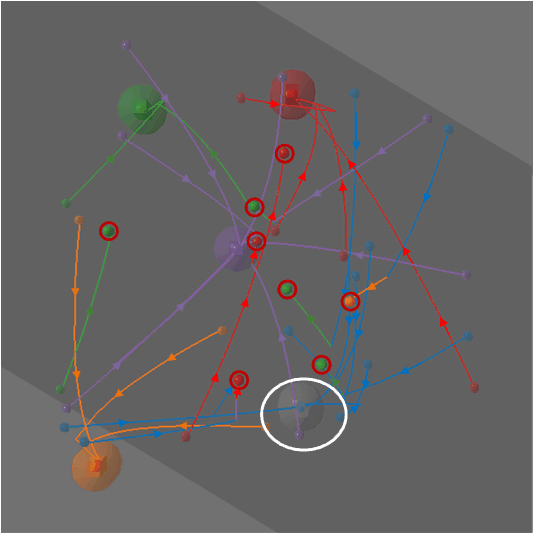}
            \caption{The greedy choice method}
        \label{rout_fig1}
    \end{subfigure}%
    \begin{subfigure}{0.24\textwidth}
        \centering
        \includegraphics[width=\linewidth]{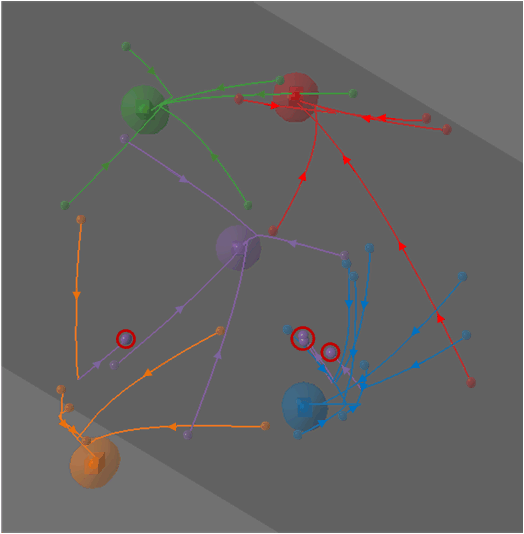}
        \caption{LIA\_MADDPG method(ours)}
        \label{rout_fig2}
    \end{subfigure}
    \caption{Compare the robot motion trajectories in a small-scale scenario after the same time step.}
    \label{rout_fig}
\end{figure}

\paragraph{\textbf{Forward-Thinking Capability }} Fig.\ref{rout_fig} visualizes the trajectories and task completion status of two different algorithms in small and medium-scale scenarios over time, where the red small circles highlight robots that have not yet completed their assigned tasks. It can be seen from  the figure  that robots following the greedy method tend to traverse more convoluted paths when compared to those employing the LIA\_MADDPG approach, which leads to higher path costs. This divergence in path complexity underscores the efficiency of LIA\_MADDPG in guiding robots to more streamlined routes.
In addition, after a period of execution,the statistical results show that over 90\% of the robots using LIA\_MADDPG have successfully completed their task assignments, while nearly thirty percent of the robots adhering to the greedy algorithm still remains  unfinished. This discrepancy demonstrates the robustness of LIA\_MADDPG in terms of task completion.

In essence, these findings suggest that LIA\_MADDPG exhibits a higher level of foresight and is more effective in yielding superior long-term results compared to strategies solely reliant on the greedy method.

\paragraph{\textbf{Collaborative Synergy Enhancement}} In Fig.\ref{rout_fig1}, the white large circle illustrates instances where multiple robots simultaneously select and endeavor to complete the same blue target task simultaneously. This concurrent pursuit can result in task exceeding the maximum robot load capacity, as indicated by it turning gray. Consequently, this may result in the robots missing out on their deserved rewards. In  contrast, LIA\_MADDPG stands out as an effective approach in coordinating actions among robots, promoting a higher level of collaboration in the task allocation process. This coordinated effort ensures that tasks are allocated efficiently and effectively, reducing the likelihood of task overload and enabling all participating robots to receive the  deserves rewards.

\section{Concluding Remarks}\label{section5}
\blue{This research has} addressed a challenging problem known as the robot swarm task allocation problem in dynamic task environment. We have modeled this problem as  Dec\_POMDP  and proposed a  novel multi-agent deep reinforcement learning  approach, called LIA\_MADDPG. In the centralized training phase, we introduce a module for local information aggregation among robots, encouraging them to focus more on information beneficial to themselves during the training process. In the distributed execution phase, we design strategy improvement methods to further enhance the quality of allocation solutions.  Finally, through extensive experiments, we have validated the effectiveness and superiority of this method in terms of convergence speed and agent cooperation performance.

It is worth pointing out that the current design of LIA\_MADDPG is most effective in large-scale homogeneous agent scenarios. It may not be well-suited for environments involving heterogeneous agents, where agents have different capabilities, tasks, and information processing requirements. Additionally, it is of interests to extend our research to incorporate collision avoidance strategies within the robot swarm task allocation framework. Therefore, a potential future direction will involve developing a multi-agent reinforcement learning approach that comprehensively considers both heterogeneous agent scenarios and collision avoidance strategies, ensuring robust performance across diverse and dynamic environments.

\bibliography{reb.bib}

\end{document}